\documentclass[sigconf,nonacm]{acmart}
\usepackage{etoolbox} %

\AtBeginDocument{%
  \providecommand\BibTeX{{%
    \normalfont B\kern-0.5em{\scshape i\kern-0.25em b}\kern-0.8em\TeX}}}

\setcopyright{acmlicensed}
\copyrightyear{2023}
\acmYear{2023}
\acmDOI{XXXXXXX.XXXXXXX}

\acmPrice{15.00}
\acmISBN{978-1-4503-XXXX-X/18/06}

\usepackage{hyperref}
\usepackage{url}

\usepackage{multicol}     %

\usepackage{amsmath}
\usepackage{graphicx}
\usepackage{subcaption}
\usepackage{listings}
\usepackage{float}
\usepackage{enumitem}
\usepackage{xspace}
\usepackage{multirow}
\usepackage{array}
\usepackage{tabularx}
\usepackage{tabu}
\usepackage{bm}

\usepackage{booktabs, makecell}
\usepackage{xcolor,colortbl}
\usepackage{cleveref}
\usepackage[english]{babel}
\usepackage{pifont} 
\usepackage{soul}
\usepackage{xfrac}
\usepackage{wrapfig}
\usepackage{mdwlist}
\usepackage{arydshln}
\usepackage{mathtools}
\usepackage{adjustbox}

\usepackage{cellspace}
\usepackage{siunitx}
\usepackage{comment}
\usepackage{microtype}

\definecolor{paleaqua}{rgb}{0.74, 0.83, 0.9}
\definecolor{aliceblue}{rgb}{0.94, 0.97, 1.0}
\newcolumntype{a}{>{\columncolor{aliceblue}}c}

\usepackage[linesnumbered,ruled,vlined,noend]{algorithm2e}
\usepackage{comment}
\definecolor{darkred}{RGB}{192, 0, 0}
\DeclareMathAlphabet\mathbfcal{OMS}{cmsy}{b}{n}

\definecolor{darkgreen}{RGB}{0,153,0}
\definecolor{dark-gray}{gray}{0.4} %

\newlist{steps}{enumerate}{1}
\setlist[steps, 1]{label = Step \arabic*:}

\newcommand{\V}[1]{\mathbf{#1}}
\newcommand{\R}{\mathbb{R}}
\newcommand{\expect}[1]{\mathrm{E}[#1]}
\newcommand{\expectlarge}[1]{\mathrm{E}\left[#1\right]}

\newcommand{\graph}{\mathcal{G}}
\newcommand{\vertexSet}{\mathcal{V}}

\newcommand{\edgeSet}{\mathcal{E}}
\newcommand{\neighbors}{\mathcal{N}_\mathcal{G}}

\newcommand{\matA}{\mathbf{A}}

\newcommand{\matW}{\mathbf{W}}

\newcommand{\matX}{\mathbf{X}}

\newcommand{\T}{\mathsf{T}}

\newcommand{\loss}{\mathcal{L}}

\newcolumntype{H}{>{\setbox0=\hbox\bgroup}c<{\egroup}@{}}

\newcommand{\method}{SuperTMA}
\newcommand{\methodrnd}{RandomTMA}
\newcommand{\methodbasic}{PSGD-PA}
\newcommand{\methodmultigpu}{GGS}

\newcommand{\methodbasicshort}{\methodbasic}

\newcommand{\llcgbasic}{LLCG}

\newcommand{\maframeworkfull}{Model Aggregation}
\newcommand{\maframework}{TMA}

\makeatletter
\def\adl@drawiv#1#2#3{%
        \hskip.5\tabcolsep
        \xleaders#3{#2.5\@tempdimb #1{1}#2.5\@tempdimb}%
                #2\z@ plus1fil minus1fil\relax
        \hskip.5\tabcolsep}
\newcommand{\cdashlinelr}[1]{%
  \noalign{\vskip\aboverulesep
           \global\let\@dashdrawstore\adl@draw
           \global\let\adl@draw\adl@drawiv}
  \cdashline{#1}
  \noalign{\global\let\adl@draw\@dashdrawstore
           \vskip\belowrulesep}}
\makeatother

\newcommand{\ignore}[1]{}

\newtheorem{theorem}{Theorem}

\renewcommand{\paragraph}[1]{\noindent\textbf{#1.}}

\begin{document}

\title{Simplifying Distributed Neural Network Training on Massive Graphs: Randomized Partitions Improve Model Aggregation}

\author{Jiong Zhu}
\authornote{This work is conducted during the authors' internship at Amazon.}
\affiliation{%
  \institution{University of Michigan}}
\email{jiongzhu@umich.edu}

\author{Aishwarya Reganti}
\affiliation{%
  \institution{Amazon}}
\email{areganti@amazon.com}

\author{Edward Huang}
\affiliation{%
  \institution{Amazon}}
\email{ewhuang@amazon.com}

\author{Charles Dickens}
\authornotemark[1]
\affiliation{%
  \institution{University of California, Santa Cruz}}
\email{cadicken@ucsc.edu}

\author{Nikhil Rao}
\affiliation{%
  \institution{Microsoft}}
\email{nikhilrao86@gmail.com}

\author{Karthik Subbian}
\affiliation{%
  \institution{Amazon}}
\email{ksubbian@amazon.com}

\author{Danai Koutra}
\affiliation{%
  \institution{Amazon \& Univeristy of Michigan}}
\email{dkoutra@{amazon.com,umich.edu}}

\begin{abstract}
Distributed training of GNNs enables learning on massive graphs (e.g., social and e-commerce networks) that exceed the storage and computational capacity of a single machine. 
To reach performance comparable to centralized training, distributed frameworks focus on maximally recovering cross-instance node dependencies with either
communication across instances or periodic fallback to centralized training,  
which create overhead and limit the framework scalability. 
In this work, we present a simplified framework for distributed GNN training that does not rely on the aforementioned costly operations, and has improved scalability, convergence speed and performance over the state-of-the-art approaches.
Specifically, our framework (1) assembles independent trainers, each of which asynchronously learns a local model on locally-available parts of the training graph, and (2) only conducts periodic (time-based) model aggregation to synchronize the local models. 
Backed by our theoretical analysis, instead of maximizing the recovery of cross-instance node dependencies---which has been considered the key behind closing the performance gap between model aggregation and centralized training---, our framework leverages randomized assignment of nodes or super-nodes (i.e., collections of original nodes) to partition the training graph such that it improves data uniformity and  
minimizes the discrepancy of gradient and loss function across instances. 
In our experiments on {social and e-commerce} networks with up to 1.3 billion edges, our proposed \methodrnd{} and \method{}
approaches---despite using \textit{less} training data---achieve state-of-the-art performance and 2.31x speedup compared to the fastest baseline,
and show better robustness to trainer failures.

\end{abstract}


\maketitle

\section{Introduction}

Graph neural networks (GNNs) achieve state-of-the-art performance on a variety of graph-based machine learning tasks with applications to recommendation systems \citep{vandenBerg17MatrixCompletion, ying18PinSage, fan19GraphRec}, fraud detection \citep{wang19FdGars, wang19ngcf, dou2020caregnn},  social network analysis \citep{qiu18deepinf, breuer20SybilEdge, cao2020coupledgnn}, and more.
As applications scale to massive social and other networks with billions of edges~\cite{zhu2019aligraph}, 
they pose scalability challenges to typical multi-layer GNN models (e.g. GCN~\citep{kipf2016semi}), 
which require a Message Flow Graph (MFG) based on each node's multi-hop neighborhood. 
These MFGs quickly exceed the storage and computational capacity of modern systems even 
under moderate batch sizes and number of GNN layers. 
This issue has motivated a productive line of work on scalable centralized GNN training
on a single instance~\citep{hamilton2017inductive,chen2018stochastic,zeng2019graphsaint,chiang2019cluster,zeng2021decoupling,fey2021gnnautoscale,narayanan2021iglu}, 
but the size of the graphs that can be trained on a single machine is ultimately limited by its available computational resources.

Distributed training overcomes the resource limitation of a single machine by leveraging parallelism on multiple machines. 
By partitioning training samples across multiple trainers and coordinating distributed updates to model weights on each trainer~\cite{narayanan2019pipedream}, data parallelism approaches
have facilitated the training of computer vision~\citep{krizhevsky2017imagenet,goyal2017accurate,yu2019parallel} and language models~\citep{mcmahan2017communication} on massive-scale datasets. 
However, \emph{graph} datasets pose additional unique challenges for data parallelism due to \emph{cross-instance node dependencies} (i.e., graph connections that reach across instance boundaries) when the data
is partitioned and distributed to multiple trainer instances. 
Different strategies have been proposed to address these challenges. %

One popular strategy, adopted by DistDGL and other frameworks~\cite{jiang2021communication,zheng2020distdgl,zheng2021distributed}, is to respect the cross-instance dependencies and 
implement communication mechanisms that allow embeddings to traverse through instance boundaries. 
To reduce the communication overhead, these approaches often distribute the training data by leveraging 
min-cut based graph partitioning algorithms (e.g.,~METIS~\citep{karypis1998fast}) and data replication. %
This strategy provides equivalency of a distributed training setup to a centralized one, 
but its reliance on excessive communication to enable unrestricted graph access across 
instances creates a bottleneck for further improving the framework speed, scalability and robustness to failures.

Another strategy is to %
initially ignore %
the cross-instance dependencies by restricting the graph access per trainer to only local graph data assigned to it, and later recover the lost data with techniques like periodic centralized training~\citep{ramezani2021learn}. 
This strategy is usually coupled with a \emph{model aggregation} mechanism, %
which periodically replaces the local model weights per trainer with aggregated weights (e.g., average) from all trainers~\citep{stich2018local,ramezani2021learn}. 
While it overcomes the overhead of excessively communicating node representations across machines, 
different implementations handle the incurred data loss and its assumed negative impact on performance 
by periodically falling back to centralized training~\citep{ramezani2021learn} or replicating nodes across  trainers~\cite{angerd2020distributed}. 
These solutions introduce new bottlenecks and additional overhead in distributed training frameworks.

\noindent \textbf{This work.} In this work, 
we revisit prior assumptions for distributed GNN training, and 
explore a simplified training setup
that removes these bottlenecks 
by discarding completely %
the cross-instance dependencies, 
and instead relies on model aggregation 
and restrictive access of each trainer to local data. 
{We focus our empirical analysis on an important graph learning task, link prediction, which (1) is the core of high-impact web-scale applications like recommendation systems~\cite{ying18PinSage}, %
and (2) is also commonly used as a pretext task for self-supervised representation learning~\cite{jin2020self} on massive graphs, %
as it learns only on edge existence when accurate node labels are unavailable or costly to obtain.} 
In this context, we discover that, surprisingly, GNN performance does \textit{not} always decrease as the number of ignored cross-instance edges increases, which suggests that other factors influence the performance.
Motivated by this observation, we re-investigate these factors, and discover through theoretical and empirical analyses that the discrepancy of data distributions among different partitions caused by min-cut partitioning algorithms 
is more critical for the reduction in performance 
than the number of ignored cross-instance edges.
To address this data discrepancy, we propose 
a simplified distributed GNN training framework using  time-based model aggregation  
along with {fast} randomized partition schemes, 
which achieves state-of-the-art performance, 
convergence speed and robustness %
without leveraging any of the aforementioned costly operations. 
We summarize our contributions as follows: 
\setlist{leftmargin=*}
\begin{itemize*}
    \item \textbf{Simplified time-based model aggregation training framework:} We present a simplified model aggregation training framework (\S\ref{sec:app-framework}) that (1)~assembles independent trainers, each of which asynchronously learns a local model on locally-available parts of the training graph, and (2)~synchronizes the local models by only conducting periodic,  time-based model aggregation that accommodates imbalanced loads and speeds among trainers. %
    
    \item \textbf{Randomized graph partition schemes with theoretical justifications:} Instead of minimizing cross-instance edges---which has been considered the key behind closing the performance gap between model aggregation and centralized training---, we discover theoretically (\S\ref{sec:partition-thms}) and empirically (\S\ref{sec:exp-benchmark}) that the discrepancy of data distributions among different partitions caused by min-cut partitioning algorithms harms the training performance. We then 
    propose improved approaches (\S\ref{sec:partition-proposed}) that partition the training graph based on {randomized assignment of nodes (which further avoids the overhead of graph clustering), or super-nodes (i.e., collections of original nodes) to trainers.}
    
    \item \textbf{Extensive empirical analysis}\footnote{Our code is available at \url{https://github.com/amazon-science/random-tma}.}:
    Our experiments, {spanning 4,600 GPU hours on 3 machines}, validate the scalability of our framework %
    on massive {social, collaboration and e-commerce networks} with up to 1.3 billion edges (\S\ref{sec:exp-setups}). We  show that our \methodrnd{} and \method{} methods---despite using less training data than the baselines (\S\ref{sec:exp-benchmark})---achieve state-of-the-art performance with a 2.31x speedup in convergence time over the fastest baseline and better robustness to trainer failures (\S\ref{sec:exp-failure}).
\end{itemize*}

\section{Related Work}
\label{sec:related}

\paragraph{Scalable GNN Training on Single Instance}
Scalable single-instance training approaches
can be grouped into three categories: 

 $\bullet$ \textit{Sampling of the Message Flow Graph} (MFG): This approach is  popular for reducing 
the complexity of message passing. %
For example, 
GraphSAGE~\cite{hamilton2017inductive} and VR-GCN~\cite{chen2018stochastic} aggregate embeddings from a subset of neighbors for each node encountered in a training step to cap the size of MFG, while ClusterGCN~\cite{chiang2019cluster}, GraphSAINT~\cite{zeng2019graphsaint} and shaDow-GNN~\cite{zeng2021decoupling}  sample a subgraph per training step and confine the training of GNN on the sampled subgraph. 

$\bullet$ \textit{Message passing pre-computation}:  
This approach relies on pre-computing aggregated features in the neighborhood of each node and using them to learn embeddings for each node independently (e.g., SIGN~\cite{frasca2020sign},  NARS~\cite{yu2020scalable}).  
However, it 
requires models like SGC \citep{wu2019sgcn} that are capable of decoupling feature aggregations from (usually linear) transformations, which restricts the GNN expressiveness. 

$\bullet$ \textit{Caching and lazy updates of stale representations or gradients}:  %
Methods in this category aim to limit the expansion of MFG. For example,  
IGLU~\citep{narayanan2021iglu} uses these techniques on backward propagation; %
GNNAutoScale~\citep{fey2021gnnautoscale} stores the historical node embeddings per layer, and only updates the stored embeddings for nodes in the mini-batch, while using the historical embeddings for the other nodes. %

Our approach on model aggregation training is orthogonal to these efforts as we focus on distributed settings with multiple trainers. Any of the approaches mentioned above can be adopted 
in our framework to further speed up each individual trainer.

\paragraph{Distributed GNN Training} 
In this work, we focus on data parallelism training on \emph{graphs}, and leave model and pipeline parallelism training~\citep{narayanan2019pipedream} as future work.
There are three key components of data parallelism that we discuss next: the scope of graph access per trainer, the data partition schemes and assignment to the trainers,
and the mechanism for model synchronization across trainers.

DistDGL~\citep{zheng2020distdgl,zheng2021distributed} and DistGNN~\citep{md2021distgnn} enable unrestricted access to the full training graph for each trainer, adopts min-cut based graph partitioning algorithms (e.g., METIS~\cite{karypis1998fast}) to partition training graph, and utilizes fully synchronous Stochastic Gradient Descent (SGD) to update local model weights after each training step; they also incorporate extensive optimization on training pipeline.
To further reduce the communication cost under a similar setup, \citet{tripathy2020reducing} optimizes matrix multiplication operations of GNNs, 
and \citet{jiang2021communication} adopts skewed sampling of MFGs to bias towards local neighbors of each node.
On the other hand, Parallel SGD with Periodic Averaging 
(PSGD-PA)~\cite{ramezani2021learn} restricts the graph access to local data only per  trainer, adopts METIS to minimize cross-partition edges ignored in training, and conducts periodic averaging to synchronize local model weights on the trainers. 
To recover more ignored cross-partition edges under this setup, LLCG~\cite{ramezani2021learn} further employs fallbacks to centralized training during the averaging process, while
\citet{angerd2020distributed}, per partition, replicates nodes from other partitions through breadth first search. 

In comparison, our proposed approach  restricts the graph access to local data only per trainer (as in PSGD-PA), but adopts randomized partition that reduces the data discrepancy across trainers and improves performance. It also employs time-based (instead of training step-based) intervals for model aggregation to accommodate heterogeneity in load and training speed on the instances. 
We compare our approaches to existing frameworks in detail in \S\ref{sec:app-comparison}.

\section{Time-based Model Aggregation \& Randomized Partition Schemes}

In this section, we first give key notation, and then present our time-based model aggregation training framework (\S\ref{sec:app-framework}). 
In \S\ref{sec:partition}, we theoretically analyze why partition schemes that minimize cross-partition edges hurt the performance of model aggregation training (\S\ref{sec:partition-thms}), and propose two randomized partition schemes %
(\S\ref{sec:partition-proposed}).
In \S\ref{sec:app-comparison}, we provide an in-depth comparison of our proposed framework to prior distributed GNN training frameworks.

\vspace{0.1cm}
\paragraph{Preliminaries} Let $\graph=(\vertexSet, \edgeSet)$ be a simple graph with node set $\vertexSet$, edge set $\edgeSet$,  adjacency matrix $\matA \in \mathbb{R}^{|\vertexSet|\times |\vertexSet|}$, and 
node feature matrix $\matX \in \R^{|\vertexSet| \times \mathcal{F}}$.
Let $\graph'=(\vertexSet',\edgeSet') \subset \graph$ be a subgraph  with $\vertexSet' \subset \vertexSet$ and $\edgeSet' \subset \edgeSet$. 
Given a node partition function $\alpha: \vertexSet \rightarrow \mathcal{I}$ on graph $\graph$ and its inverse $\alpha^{-1}(i) = \{v \ \vert\ v \in \vertexSet \land \alpha(v) = i\}$, we define the node-induced subgraph $\graph^{(i)} = (\vertexSet^{(i)}, \edgeSet^{(i)})$ of partition $i \in \mathcal{I}$ as $\vertexSet^{(i)} = \alpha^{-1}(i)$ and $\edgeSet^{(i)} = \{(u, v)\ |\ (u, v) \in \edgeSet \land u, v \in \alpha^{-1}(i)\}$. 
For a graph with node class labels $y_v \in \mathcal{Y}$, %
we define its homophily ratio as the fraction of homophilous edges  linking same-class nodes~\citep{zhu2020beyond}: $h = |\{(u,v)\ |\ (u,v) \in \edgeSet \land y_u = y_v\}|/|\edgeSet|$.
We refer to a graph with $h\geq 0.5$ as homophilic.

\vspace{0.1cm}
\paragraph{Problem Statement} %
Given a large-scale, simple graph $\graph=(\vertexSet,\edgeSet)$ 
and its node feature matrix $\matX$, 
we aim to efficiently train a GNN model in a distributed setting with $M$ trainers  
in order to learn weight matrix $\matW$ 
of GNN model $g(u, v, \matX, \graph; \matW)$ to effectively predict the probability of edge existence  $(u, v) \notin \edgeSet$. 
{We have two requirements to address the bottlenecks of prior work:}
(\texttt{R1})~a fast and effective partition function $\alpha: \vertexSet \rightarrow \mathcal{I}$ that maps portions of the graph to the trainers; and (\texttt{R2})~a low-cost way of combining the learned parameters of different trainers in order to achieve high prediction performance.

\begin{figure}[t]
    \centering
    \includegraphics[width=\columnwidth]{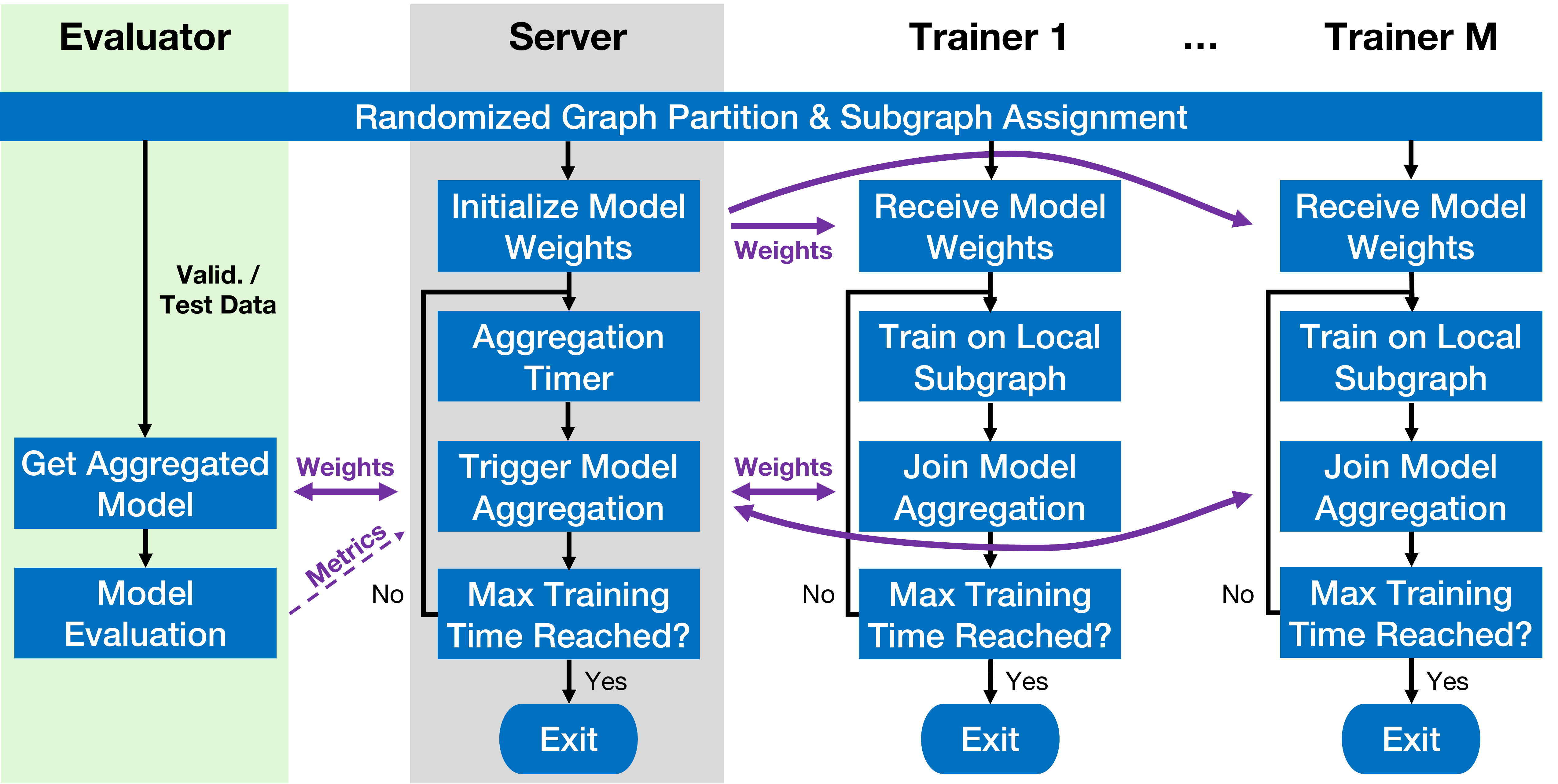}
    \caption{Architecture of our Time-based Model Aggregation (\maframework{}) framework. We provide the pseudo-code for the server and trainer in Alg.~\ref{alg:ama-server} and \ref{alg:ama-client}, respectively. We use solid purple arrows to represent synchronous communications, and dashed ones for asynchronous communications.
    }
    \label{fig:system-arch}
\end{figure}

\begin{algorithm}[t]
    \DontPrintSemicolon 
    \SetAlgoLined
    \SetKwInput{KwInput}{Input}                %
    \SetKwInput{KwOutput}{Output}  
    \SetAlgoLined
    \caption{Time-based Model Aggregation Server}
    \label{alg:ama-server}
    \KwInput{Total training time $\Delta T_{\mathrm{train}}$, model aggregation interval $\Delta T_{\mathrm{int}}$ \& operator $\phi$, validation \& test sample splits and subgraphs $\graph_{\mathrm{val}}$ \& $\graph_{\mathrm{test}}$, trainer IDs $\{1,\cdots,M\}$ and  network addresses, \texttt{optimizer} function, model configurations \& hyperparameters.} 
    Establish communication with trainers and distributed Key-Value Store \texttt{KV}; broadcast \texttt{optimizer} function, model configurations \& hyperparameters.\;
    Setup model and initialize model weights $\matW_{\mathrm{global}}[0]$.\;
    Wait until \texttt{all}([$\texttt{KV[ready][i]}$ for $i \in \{1,\cdots,M\}$]). \;
    \texttt{KV[agg]} = \texttt{KV[stop]} = \texttt{False} \;
    Broadcast initialized model weights $\matW_{\mathrm{global}}[0]$ to trainers. \;
    $T_{\mathrm{start}} = T_{\mathrm{agg}} = \texttt{current\_time()}$; $t = 0$. \;
    \While{not \texttt{KV[stop]}}{
        \If{$\texttt{current\_time()} - T_{\mathrm{agg}} \geq \Delta T_{\mathrm{int}}$}
        {
            \texttt{KV[agg]} = \texttt{True}\;
            Receive weights $\matW_i$[t] from trainer $i\in \{1, \cdots,M\}$.\;
            \texttt{KV[agg]} = \texttt{False}\;
            $\matW_{\mathrm{global}}[t+1] = \phi(\matW_1[t], \cdots, \matW_M[t])$.\;
            Broadcast global weights $\matW_{\mathrm{global}}[t+1]$ to trainers. \;
            Invoke \texttt{metrics}[t+1] = \texttt{eval}($\matW_{\mathrm{global}}[t+1]$, $\graph_{\mathrm{val}}$) on an evaluation process.\;
            $t = t + 1$\;
        }
        \If{$\texttt{current\_time()} - T_{\mathrm{start}} > \Delta T_{\mathrm{train}}$}
        {
            \texttt{KV[stop]} = \texttt{True}\;
        }
    }
    Wait until \texttt{metrics}[t] is ready; 
    $t_* = \texttt{arg\_best}(\texttt{metrics})$. \;
    \texttt{metrics}[$t_*$] = \texttt{eval}($\matW_{\mathrm{global}}[t_*]$, $\graph_{\mathrm{test}}$).\;
    \KwOutput{Best model weights $\matW_{\mathrm{global}}[t_*]$, \texttt{metrics} and $t_*$.}
\end{algorithm}

\begin{algorithm}[t]
    \DontPrintSemicolon 
    \SetAlgoLined
    \SetKwInput{KwInput}{Input}                %
    \SetKwInput{KwOutput}{Output}  
    \SetAlgoLined
    \caption{Time-based Model Aggregation Trainer}
    \label{alg:ama-client}
    \KwInput{Trainer ID $i \in {1,\cdots,M}$; \\ $\;\;\;\;\;\;\;\;\;\;$ assigned training subgraph $\graph_{\mathrm{train}}^{(i)} \subset \graph_{\mathrm{train}}$.}
    Establish communication with the server and distributed Key-Value Store \texttt{KV}; receive \texttt{optimizer} function, model configurations \& hyperparameters.\; Initialize $\texttt{KV[ready][i]} = \texttt{False}$.\;
    Load $\graph_{\mathrm{train}}^{(i)}$; prepare data for training; set up GNN model.\;
    $\texttt{KV[ready][i]} = \texttt{True}$\;
    Receive initialized model weights $\matW_{\mathrm{global}}[0]$ from server.\;
    $t$ = 0 \; %
    \While{not \texttt{KV[stop]}}{
        Construct mini-batch $\xi_{i}^{(t)}$ on local subgraph $\graph_{\mathrm{train}}^{(i)}$.\;
        $\matW_i[t] = \texttt{optimizer}(\xi_{i}^{(t)}, \matW_i[t])$ \;
        \If{\texttt{KV[agg]}}{
            Send local model weights $\matW_i$[t] to the server.\;
            Get global model weights $\matW_{\mathrm{global}}[t+1]$ from server.\;
            Overwrite local weights $\matW_i$[t]$\leftarrow$ $\matW_{\mathrm{global}}[t+1]$.\;
            $t = t + 1$\;
        }
    }
\end{algorithm}

\subsection{\maframework{}: Proposed Time-based Model Aggregation Framework} 
\label{sec:app-framework}

To solve the above-mentioned problem, we introduce a simplified distributed GNN framework that leverages the idea of Time-based Model Aggregation (\maframework{}). 
Figure~\ref{fig:system-arch} illustrates the architecture of our \maframework{} framework, which  
consists of $M$ trainer processes, a server process, and one or more evaluation processes.
These processes may run on a cluster of machines or a single machine based on the scale of the dataset and availability of resources. 
The design of the server and trainer processes are formally presented in Algorithms~\ref{alg:ama-server} and \ref{alg:ama-client}, respectively.

Each \textit{trainer} process $i \in \{1,\cdots,M\}$ loads a part of the training graph $\graph_{\mathrm{train}}^{(i)} \subset \graph_{\mathrm{train}}$, and conducts stochastic gradient descent on mini-batches sampled \textit{solely} from the local training subgraph $\graph_{\mathrm{train}}^{(i)}$ assigned to it via partition function $\alpha$.  
We discuss partition options and propose improved approaches for partitioning and assigning the local training subgraphs (requirement \texttt{R1}) in~\S\ref{sec:partition}. %

On the \textit{server} side, to satisfy requirement \texttt{R2} in our problem statement, our TMA framework periodically executes a model aggregation operation $\phi$ to synchronize the learned model parameters $\matW_i$ across trainers.  
This procedure is triggered on a time-based interval, supporting asynchronous training across heterogeneous trainers;  
this is critical for a scalable and efficient framework as we empirically observe (in \S\ref{sec:exp-efficiency})
that the number of training steps finished on the slowest trainer can be up to 28.8\% less than that on the fastest trainer.
{For the choice of aggregation operator}, we find that simply averaging the model parameters of the trainers provides better performance over more complex model aggregation operators that consider the loss on different trainers.
{We use separate \textit{evaluator} processes that run along side the server and training processes to evaluate the aggregated model.}

\subsection{Improving Time-based Model Aggregation with Randomized Partition Schemes}
\label{sec:partition}
Before proposing our improved partition schemes that satisfy requirement \texttt{R1}, we review the role of partitioning in distributed GNN training and highlight key limitations of prior work.

Graph partitioning and assignment is a standard preprocessing step for distributed GNN training: the full training graph $\graph_{\mathrm{train}}$ is first partitioned into smaller subgraphs $\graph_{\mathrm{train}}^{(i)} \subset \graph_{\mathrm{train}}$, which are then assigned to different trainers $i \in \{1,\cdots,M\}$. 
Existing frameworks, such as DistDGL, \methodbasic{} and \llcgbasic{}, focus on maximizing the coverage of cross-instance node dependencies in their partition schemes.
Specifically, they generate $M$ partitions of the training graph such that they minimize the %
edge cuts (e.g., with METIS~\citep{karypis1998fast}), and then map them one-to-one to the $M$ trainers.
For model aggregation-based frameworks, %
recovering the cross-partition edges (through optimal partitions and other mechanisms) has also been considered the key behind minimizing the performance gap between distributed and centralized GNN training~\citep{ramezani2021learn}. 
However, our analysis reveals that partitions minimizing the number of cross-machine edges
also lead to increased disparity of training data across different trainers, which in turn leads to discrepancy of gradients and training losses that stall the convergence of model aggregation training (\S\ref{sec:partition-thms}).\footnote{{Although METIS supports balancing nodes with different labels across partitions, this is incompatible with our focus on link prediction as (1)~this task does not use node labels during training; and %
(2)~obtaining accurate node labels can be costly for web-scale applications.  Even in node classification, only a small portion of labels is available during training.}}
To mitigate the disparity between partitions, 
we propose randomized partition schemes at the node or super-node level  (\S\ref{sec:partition-proposed}), which achieve  improved performance and convergence speed despite using less training data due to discarding the cross-instance edges %
(\S\ref{sec:exp-benchmark}).

\subsubsection{Minimizing cross-partition edges harms model aggregation training.}
\label{sec:partition-thms}
The residual error of the loss function and its gradients caused by the local-access constraint is considered the key behind the performance gap between model aggregation training and centralized training~\cite{ramezani2021learn}; in other words, the mismatch of the loss values and their gradients on different distributed trainers and to those of a centralized trainer hurts the performance of model aggregation training.   
Here we provide a theoretical analysis about how the popular approach in existing distributed frameworks~\cite{zheng2020distdgl,zheng2021distributed,md2021distgnn,ramezani2021learn,angerd2020distributed} of one-to-one mapping of METIS partitions to trainers, which minimizes the cross-partition edges, contributes to the residual error in the gradient descent process of model aggregation training on homophilic graphs. %

{We analyze a case of a homophilic graph, where the disparity of partitions is measured by the difference of the feature distributions, which correlate with two class labels.}
In Lem.~\ref{thm:homophily-min-cut}, we show that partitions minimizing the number of cross-partition edges amplify the differences of feature distributions among partitions, which in the case we assume leads to complete separation of nodes from different classes. %

\begin{lemma}
    \label{thm:homophily-min-cut}
    Assume a homophilic graph with two equally-sized classes 
    and edges modeled by a class compatibility matrix $\mathbf{H}$ \cite{zhu2020beyond} as follows:
    the probability $p_{ji}$ of node $j$ linking to node $i$ satisfies
    $$p_{ji} \propto \mathbf{H}(y_i, y_j)=
    \begin{cases}
    h \geq 0.5, & \text{for } y_i = y_j \\
    1-h & \text{otherwise,}
  \end{cases}
  $$
  where $y_i, y_j$$\in$$\{0, 1\}$ are the class labels of nodes $i,j$.
  Let $\V{x}_v = \mathrm{onehot}(y_v)$ be features of node $v \in \vertexSet$, and $\alpha: \vertexSet \rightarrow \{1, 2\}$ be a function that assigns the nodes into two equally-sized partitions. Then, the smallest expected edge-cut is reached when each partition has same-class nodes with the same features: $\alpha(i) = \alpha(j)$ iff $y_i = y_j$ or $\V{x}_i = \V{x}_j$.
\end{lemma}

In Thm.~\ref{thm:homophily-grad-loss}, we demonstrate the effects of disparity between partitions by showing that it leads to discrepancy between (initial) gradients and loss derived on different trainers, which is the key factor affecting the performance under model aggregation training when only local data is used~\cite{ramezani2021learn}. 

\begin{theorem}
    \label{thm:homophily-grad-loss}
    Given the same homophilic graph with two classes $y\in \{0, 1\}$ and partition function $\alpha: \vertexSet \rightarrow \{1, 2\}$ as in Lem.~\ref{thm:homophily-min-cut}, suppose the feature distribution of each partition is $\mathbf{C}_{1}, \mathbf{C}_2 \in [0, 1]^2$, respectively.
    Consider a 1-layer GNN formulated as $\V{z}=f(\matA,\matX)=\sigma(\bar{\matA}\matX\matW$) for node classification, with row-normalized adjacency matrix $\bar{\matA}$, sigmoid function $\sigma$, node features $\V{x}_v = \mathrm{onehot}(y_v)$, and a L2-loss function 
    $\mathcal{L}(y, z) = \tfrac{1}{2}\Vert \V{y} - \V{z} \Vert^2$ for training. Then, we have:
    \begin{enumerate*}
        \item When initializing $\matW=\mathbf{0}$, the discrepancies among the expected initial local gradients $\expect{\nabla\loss_{i}^{local}}, i\in\{1,2\}$ on each instance, without considering cross-partition edges, and the expected initial gradient $\expect{\nabla\loss^{global}}$ for centralized training increase with the differences of the group distributions $\Vert \mathbf{C}_{2} - \mathbf{C}_{1} \Vert$.
        
        \item For arbitrary learned model weights $\matW$, the expected loss values $\expect{\loss_{i}^{local}(\matW)}$ on each instance $i\in\{1,2\}$, without considering cross-partition edges, is equal if and only if  $\,\mathbf{C}_{1} = \mathbf{C}_{2}$.
    \end{enumerate*}
    
\end{theorem}

We give the proofs of both Lem.~\ref{thm:homophily-min-cut} and Thm.~\ref{thm:homophily-grad-loss} in App. \S\ref{app:proofs}. 
While our theoretical analysis holds under specific assumptions, we discuss the empirical observations on the discrepancy of loss functions among different trainers under more generalized settings on real-world datasets in \S\ref{sec:exp-benchmark}.

\subsubsection{Proposed Randomized Graph Partition}
\label{sec:partition-proposed}

Based on our analysis that disparity of training graph partitions stalls the convergence under model aggregation training, we propose two simple but effective randomized partition schemes that reduce this disparity in model aggregation training, and combine them with our time-based training framework:
\methodrnd{} leverages randomized partition of nodes, and \method{} leverages randomized partition of super-nodes (i.e., collections of nodes)~\cite{LiuSDK18}. 

\vspace{0.1cm}
\paragraph{\methodrnd{}: Randomized Node Partition-based \maframework{}} %
The idea of randomized node partition is simple: 
each node is randomly and independently assigned to one of the graph partitions, and 
the node-induced subgraph $\graph^{(i)}$ of each partition $i$ is assigned 
to the trainers through an one-to-one mapping. %
Since the assignment of each node is considered independently, this partition scheme does not bias towards minimizing the cross-partition edges:  
the probability of each edge that does not connect nodes in different partitions is $\tfrac{1}{M}$, where $M$ is the number of trainers. 
Despite having less data available for model aggregation training than clustering-based frameworks, this partition scheme eliminates the time and cost of graph clustering (c.f. Table~\ref{tab:exp-ablation-base-models}), and the expected disparity of training data on different partitions. We formalize the latter next: %
\begin{corollary}
    Given the same homophilic graph with two class labels $y\in \{0, 1\}$ and partition function $\alpha: \vertexSet \rightarrow \{1, 2\}$ as in Lem.~\ref{thm:homophily-min-cut}, when the nodes  are randomly assigned to each partition under independent and identical distributions, the following hold:
    \vspace{-0.15cm}
    \begin{enumerate*}
        \item $\expect{\V{C}_1 - \V{C}_0} = \V{0}$.
        \item For training the GNN described in Thm.~\ref{thm:homophily-grad-loss}, the expected loss values $\expect{\loss_{i}^{local}(\matW)}$ and gradients $\expect{\nabla\loss_{i}^{local}}$ are equal across trainers $i \in \{0, 1\}$ 
        for arbitrary model weights $\matW$.
    \end{enumerate*}
\end{corollary}

We demonstrate the generalizability of this corollary on real-world datasets and different learning tasks in \S\ref{sec:exp-benchmark}. Specifically, we observe that \methodrnd{}
reduces the differences in loss functions across different trainers, achieves comparable or better performance than existing distributed training approaches, and has faster convergence speed despite using significantly less training data than frameworks  that rely on min-cut partitioning.  

\vspace{0.1cm}
\paragraph{\method{}: Randomized Super-Node Partition-based \maframework{}} %
This partition scheme combines (1) the ability of node-level randomized partition in \methodrnd{} to handle the data disparity issue 
with (2) the better training data availability and robustness to overfitting of clustering-based partitions (as in \methodbasic{} and \llcgbasic{}~\cite{ramezani2021learn}).
At a high level, it randomly assigns super-nodes or mini-clusters\footnote{Similar to our work, ClusterGCN~\cite{chiang2019cluster} also leverages mini-clusters but it does so in order to form mini-batches for scalable single-instance training; on the other hand, we use mini-clusters to partition the graph for distributed training.} generated by clustering algorithms to each partition. 
Specifically, we first use an efficient clustering algorithm like METIS to generate $N \gg M$ mini-clusters for training on $M$ instances.
Each mini-cluster is treated as a 
super-node  
and is randomly assigned to a graph partition similar to \methodrnd{}. 
Then,  training subgraph $\graph^{(i)}_{train}$ is derived as the subgraph induced by all the collections of  \emph{nodes} assigned to  partition $i$ (i.e., the union of the nodes in all its assigned super-nodes). 

The use of super-nodes generated by clustering algorithms reduces the loss of cross-partition edges compared to \methodrnd{}, which mitigates the issue of overfitting 
{on smaller datasets or smaller graph partitions when using a large number of trainers. In both cases,  \method{} shows better performance than \methodrnd{} and benefits more from an increased number of trainers (\S\ref{sec:exp-benchmark}, \S\ref{sec:exp-ablation-num-trainers})}.
By adjusting the number of super-nodes $N$, we can control the trade-off between minimizing the cross-partition edges and the data disparity among partitions: when $N = M$ (i.e., the number of super-nodes equals to the number of trainers), we minimize the cross-partition edges (similar to previous approaches).
Note that \methodrnd{} is equivalent to \method{} when $N = |\vertexSet|$, but without the overhead of graph clustering. %

\subsection{Comparison with Existing Frameworks} %
\label{sec:app-comparison}
\subsubsection{\maframework{} vs.\ DistDGL}
DistDGL~\citep{zheng2021distributed} assumes that each trainer (or mini-batch sampler) has access to the full training graph $\graph_{\mathrm{train}}$; whereas our Time-based \maframeworkfull{} (\maframework) framework only allows each trainer $i$ to access its local training subgraph $\graph_{\mathrm{train}}^{(i)} \subset \graph_{\mathrm{train}}$.
The more restrictive access to the training data in the \maframework{} framework reduces the amount of available training samples and is widely believed to result in inferior performance in previous works~\cite{ramezani2021learn}. 
However, we show in \S\ref{sec:partition} that with our proposed partition schemes, which minimize the discrepancy of gradient and loss function across trainers, the \maframework{} framework can achieve better or comparable performance to DistDGL with improved convergence speed.
In addition, DistDGL uses synchronous SGD, which requires the gradients of trainers to be synchronized after each training step; 
\maframework{} only periodically synchronizes the model weights (instead of gradients) among trainers, which significantly reduces the number of synchronizations and allows asynchronous training steps before time-based model aggregation. 

\subsubsection{\maframework{} vs.\ \methodbasic{} and \llcgbasic}
While the \methodbasic{} and \llcgbasic{} approaches~\cite{ramezani2021learn} are also designed upon the model aggregation mechanism, they adopt  a different approach to mitigate the performance gap compared to global-access and fully-synchronous approach like DistDGL: \methodbasic{} uses one-to-one mapping of METIS clusters to trainers to minimize the number of cross-partition edges, and \llcgbasic{} further employs periodical fallbacks to centralized training to recover more cross-instance edges. 
In contrast, our \maframework{} framework discards the cross-instance dependencies (resulting in significantly fewer training edges) and  leverages randomized partition schemes that reduce the disparity of training data among trainers. %
Also, the design of \methodbasic{} and \llcgbasic{}  requires more synchronization of the training progress across different trainers, as %
averaging 
is triggered after a certain number of training steps per trainer, %
while our framework utilizes time-based aggregation intervals to accommodate different training speeds across instances.

\section{Empirical Analysis}
\label{sec:exp}

In this section, we seek to address the following research questions: 
\textbf{(Q1)}~How does the convergence speed and performance of the proposed approaches, \methodrnd{} and \method{}, compare with other training approaches? 
{\textbf{(Q2)}~What factors contribute to the improved convergence speed and performance of \methodrnd{} and \method{} over the baselines?}
\textbf{(Q3)}~How robust are \methodrnd{} and \method{} to different hyperparameters, such as 
aggregation interval and number of trainers?
\textbf{(Q4)}~Are the performance and convergence time of \methodrnd{} and \method{} robust to possible failure of trainers in a distributed setting? 

\subsection{Experimental Setup}
\label{sec:exp-setups}

\begin{table}[t]
    \caption{Dataset statistics.}
    \label{table:dataset-stats}
    \vspace{-0.2cm}
    \resizebox{\columnwidth}{!}
    {
    \begin{tabular}{l r  r  r  c  H}
        \toprule
        \textbf{Dataset} & \textbf{\#Nodes} $|\vertexSet|$ & \textbf{\#Edges} $|\edgeSet|$ & \textbf{\#Feat.} $F$ & \textbf{\#Val. / Test Edges} & \textbf{Feat. Sim.} $h_F$ \\
        \midrule
        \textbf{Reddit} & 232,965 & 114,615,892 & 602 & 114,615 / 114,617 & 0.124 \\ %
        \textbf{ogbl-citation2} & 2,927,963 & 30,561,187 & 128 & 86,956 / 86,956 & 0.855 \\ %
        \textbf{MAG240M-P} & 121,751,666 & 1,297,748,926 & 768 & 122,088 / 129,781 & 0.005 \\
        \midrule
        \textbf{E-comm} & 33,886,911 & 207,157,590 & 300 & 1,232,708 / 123,270,705 \\
        \bottomrule 
    \end{tabular}
    }
    \vspace{-0.5cm}
\end{table}

\paragraph{Dataset and evaluation setup}
{We focus on the important graph learning task of link prediction due to its close tie with recommendation systems~\cite{ying18PinSage} and self-supervised representation learning~\cite{jin2020self}. We consider four large-scale networks for our experiments}: 
(1)~Reddit~\citep{hamilton2017inductive}, a network connecting posts with common commenters; 
(2)~ogbl-citation2~\citep{hu2020ogb}, a paper citation network;
(3)~MAG240M-P, the paper citation network extracted from MAG240M~\citep{hu2021ogblsc}, and,
{(4)~E-comm, a proprietary e-commerce dataset of queries and items, which are sampled from anonymized logs of four different market locales of this e-commerce store, with edges connecting items to related items
and queries to related items, and node features generated by a fine-tuned BERT model.}
We list the statistics of these datasets in Table~\ref{table:dataset-stats}. To our knowledge, MAG240M-P is the largest publicly-available homogeneous benchmark network  with 768-dimensional node features (175GB in storage) and over 1.29 billion edges. %
For ogbl-citation2, we use the train / validation / test splits provided with the dataset; 
for Reddit and MAG240M-P which are originally proposed as node classification benchmarks, we create the validation / test splits by randomly selecting and removing one outgoing edge per node in the validation / test splits of node classification;
{for E-comm, we use all item correlations and 3 months of query-item associations for training, and use the next month of query-item associations for model evaluation.}
We report the Mean Reciprocal Rank (MRR) of the predicted score of each positive candidate in validation / test splits over 1,000 randomly selected negative candidates, which are fixed across different experiment runs.
We do not use neighborhood sampling in the evaluation process as it introduces additional randomness to the test results. 

\paragraph{Hardware Specifications} {We spend $\sim$4,600 GPU hours on a maximum of three AWS EC2 p3.16xlarge instances for our experiments}, with each instance featuring 64 CPU cores, 488 GB RAM, and 8 NVIDIA Tesla V100 GPU with 16 GB Memory per GPU. 

\paragraph{Trainer setup} {To keep the empirical analysis resource- and cost-efficient}, we run experiments with $M=3$ training processes on Reddit and ogbl-citation2 on a single physical instance; for MAG240M-P, we run $M=3$ training processes on 2 physical instances by default. In \S\ref{sec:exp-ablation-num-trainers}, {we further report the results of $M=5$ and the maximum $M=23$ training processes using all 3 physical instances. We give the distribution of the trainers on physical instances in App.~\ref{app:exp-details}. }

\paragraph{Training Approaches} We compare the convergence speed and performance for two variants of our proposed training approach, \methodrnd{} and \method{} (with number of super-nodes $N=15,000$), along with the following baselines: 
(1)~\methodbasic{}~\citep{ramezani2021learn}, which {we implemented as} a special case of \method{} with the number of super-nodes $N$ equal to the number of trainers $M$ to minimize the cross-machine edges; {while its original design conducts synchronization on a step-based interval, we enhance it with our time-based model aggregation mechanism and focus our analysis on the effects of its partition scheme.}
(2)~Learn locally, correct globally (LLCG) \citep{ramezani2021learn}, which behaves similarly to \methodbasic{} {(we also enhance it with our time-based model aggregation mechanism in our experiments)}, but has an additional step of global model correction on the server in the model aggregation process; 
(3)~Global Graph Sampling (GGS)~\cite{ramezani2021learn,zheng2021distributed,md2021distgnn}, where each trainer has unrestricted access to the full training graph, with local models on trainers updated through synchronous SGD {to synchronize the gradients among trainers after each training step}. We implement GGS using the MultiGPU training functionality, 
where each trainer runs on a separate GPU of the physical machine; 
though DistDGL~\citep{zheng2021distributed} is not directly compatible with our implementation, our implementation emulates its training pipeline {and represents an ideal version of DistDGL without the communication overhead of accessing node embeddings remotely}. 
For all approaches, we create a separate process for model evaluation as in Fig.~\ref{fig:system-arch}, and adopt the same interval for model evaluation to ensure a fair comparison.

\paragraph{GNN Encoders} We consider two GNN choices for encoders {on homogeneous graphs}: GCN~\citep{kipf2016semi} and GraphSAGE~\citep{hamilton2017inductive}. In addition, we adopt MLP as an additional baseline, as previous works have revealed that GNNs are not guaranteed to perform better than a graph-agnostic baseline~\citep{zhu2020beyond}.
For the heterogeneous E-comm dataset, we test GCN~\citep{kipf2016semi} and RGCN~\citep{schlichtkrull2018modeling} as encoders.
For all models, we follow \citet{chen2018stochastic} and \citet{you2020design} and use PReLU as non-linear activation function, and Layer Normalization~\citep{ba2016layer} before activation to improve performance of all encoders.
we list more hyperparameters for encoder in App.~\S\ref{app:exp-details}.

\paragraph{Link Prediction Decoder} 
On homogeneous graphs, we use an MLP decoder to predict the link probability
between a pair of nodes: we find in our experiments that multi-layer MLP with non-linearity significantly improves the link prediction performance over a vanilla dot product decoder. 
{We elaborate on the formulation of MLP decoder in App.~\S\ref{app:exp-details}.
We additionally test DistMult~\citep{yang2015embedding} as a decoder for heterogeneous graphs on E-comm dataset.
To ensure a fair comparison between different training approaches and encoders, we fix on each dataset the number of layers and the sizes of hidden states for the decoder; we list these parameters in App.~\S\ref{app:exp-details}.}

\paragraph{Mini-batch and Negative Sampling}
For all trainings, we randomly select edges in the training set to form the mini-batches, and use GraphSAGE sampler~\cite{hamilton2017inductive} to reduce the size of Message Flow Graph (MFG). For each positive edge sample $(u, v)$ in the mini-batches, we randomly sample one edge $(u, v')$ with a different tail $v' \in \vertexSet$ as the negative sample.

\vspace{-0.3cm}
\subsection{(Q1) Performance and Convergence Speed}
\label{sec:exp-benchmark}

\begin{table*}[t]
    \centering
    \caption{Comparison of different training approaches on link prediction: ratio $r$ of training samples (i.e., edges in the training graph) available to each approach, performance (Test MRR), convergence time (in minutes). 
    We report the performance for each training approach as the test MRR obtained on the best encoder (more details in Table~\ref{tab:exp-ablation-base-models} and \ref{tab:exp-ablation-base-models-ecomm}), and the convergence time as the time to reach within 1\% interval relative to its maximum validation MRR. 
    The average rank is calculated as the average rankings of MRR and convergence time for each approach across all datasets. 
    Despite having fewer training samples, the proposed approaches, \method{} and \methodrnd{}, perform contrary to the expectation of prior works and achieve the best performance with the highest average ranks in MRR. The faster variant, \methodrnd{}, has the best convergence speed overall with up to 2.31x speed up to the fastest baseline, while still achieving comparable performance to \method{}. 
    }
    \footnotesize 
    \label{tab:exp-benchmark}
    \vspace{-0.2cm}
    \resizebox{\textwidth}{!}{
        \begin{tabular}{H H l cH ccc c ccc c ccc c ccc c aa}
        \toprule
        
        \multirow{3}{*}{\textbf{Sampler}} 
        & \multirow{3}{*}{\textbf{\shortstack[l]{GNN}}} 
        & \multirow{3.5}{*}{\textbf{\shortstack[l]{Training\\Approach}}}
        & \multirow{3.5}{*}{\textbf{\shortstack[c]{\#Parts\\($N$)}}}
        & 
        & \multicolumn{3}{c}{\multirow{1}{*}{\shortstack[c]{\textbf{Reddit} \textcolor{gray}{($|\edgeSet|$ = 114M)}}}} &
        & \multicolumn{3}{c}{\multirow{1}{*}{\shortstack[c]{\textbf{citation2} \textcolor{gray}{($|\edgeSet|$ = 30.5M)}}}} &
        & \multicolumn{3}{c}{\multirow{1}{*}{\shortstack[c]{\textbf{MAG240M-P} \textcolor{gray}{($|\edgeSet|$ = 1.30B)}}}} &
        & \multicolumn{3}{c}{\multirow{1}{*}{\shortstack[c]{\textbf{E-comm} \textcolor{gray}{($|\edgeSet|$ = 207M)}}}} &
        & \multicolumn{2}{a}{\multirow{1}{*}{\shortstack[c]{\textbf{Average Rank}}}}
        \\
        \cmidrule{6-8} \cmidrule{10-12} \cmidrule{14-16} \cmidrule{18-20} \cmidrule{22-23}
        &&&&  
        & \shortstack[c]{\textbf{Ratio}\\\textcolor{white}{(}$r$\textcolor{white}{)}} & \multicolumn{1}{c}{\shortstack[c]{\textbf{MRR}\\(\%)}} & \multicolumn{1}{c}{\shortstack[c]{\textbf{Time}\\(min)}} &
        & \multicolumn{1}{c}{\shortstack[c]{\textbf{Ratio}\\\textcolor{white}{(}$r$\textcolor{white}{)}}} & \multicolumn{1}{c}{\shortstack[c]{\textbf{MRR}\\(\%)}} & \multicolumn{1}{c}{\shortstack[c]{\textbf{Time}\\(min)}} &
        & \multicolumn{1}{c}{\shortstack[c]{\textbf{Ratio}\\\textcolor{white}{(}$r$\textcolor{white}{)}}} & \multicolumn{1}{c}{\shortstack[c]{\textbf{MRR}\\(\%)}} & \multicolumn{1}{c}{\shortstack[c]{\textbf{Time}\\(min)}} & 
        & \multicolumn{1}{c}{\shortstack[c]{\textbf{Ratio}\\\textcolor{white}{(}$r$\textcolor{white}{)}}} & \multicolumn{1}{c}{\shortstack[c]{\textbf{MRR}\\(\%)}} & \multicolumn{1}{c}{\shortstack[c]{\textbf{Time}\\(min)}} & 
        & \multirow{1}{*}{{\bf MRR\ }} & \multirow[t]{1}{*}{{\bf Time}} \\
        
        \midrule

        \multirow{5}{*}{SAGE} & \multirow{5}{*}{GCN} & \textbf{\methodrnd} & $|\vertexSet|$ & & 0.33 & $\underset{\scriptstyle{\pm \textrm{0.21}}}{\textrm{47.78}}$ & $\underset{\scriptstyle{\pm \textrm{7.1}}}{\textrm{67.4}}$ &  & 0.33 & $\underset{\scriptstyle{\pm \textrm{0.24}}}{\textrm{83.28}}$ & $\underset{\scriptstyle{\pm \textrm{14.3}}}{\textrm{56.4}}$\cellcolor{green!20} &  & 0.33 & $\underset{\scriptstyle{\pm \textrm{0.09}}}{\textrm{85.77}}$\cellcolor{blue!20} & $\underset{\scriptstyle{\pm \textrm{27.6}}}{\textrm{169.3}}$\cellcolor{green!20} &  & 0.33 & $\underset{\scriptstyle{\pm \textrm{0.02}}}{\textrm{84.12}}$ & $\underset{\scriptstyle{\pm \textrm{20.0}}}{\textrm{52.5}}$\cellcolor{green!20} &  & {2.0} & \underline{\bf 1.5} \\
        & & \textbf{\method} & 15,000 & & 0.35 & $\underset{\scriptstyle{\pm \textrm{0.64}}}{\textrm{48.68}}$\cellcolor{blue!20} & $\underset{\scriptstyle{\pm \textrm{6.9}}}{\textrm{154.4}}$ &  & 0.58 & $\underset{\scriptstyle{\pm \textrm{0.43}}}{\textrm{83.75}}$\cellcolor{blue!20} & $\underset{\scriptstyle{\pm \textrm{39.6}}}{\textrm{126.8}}$ &  & 0.64 & $\underset{\scriptstyle{\pm \textrm{0.36}}}{\textrm{85.27}}$ & $\underset{\scriptstyle{\pm \textrm{0.1}}}{\textrm{189.5}}$ &  & 0.76 & $\underset{\scriptstyle{\pm \textrm{0.45}}}{\textrm{84.44}}$\cellcolor{blue!20} & $\underset{\scriptstyle{\pm \textrm{12.8}}}{\textrm{126.3}}$ &  & \underline{\bf 1.2} & {3.5} \\
        
        \noalign{\vskip 0.25ex}
        \cdashline{3-23}[0.8pt/2pt]
        \noalign{\vskip 0.25ex}
        
        & & \cellcolor{gray!20}\textbf{\methodbasic} & $M=3$ & & 0.88 & $\underset{\scriptstyle{\pm \textrm{0.35}}}{\textrm{46.02}}$ & $\underset{\scriptstyle{\pm \textrm{10.0}}}{\textrm{37.2}}$\cellcolor{green!20} &  & 0.95 & $\underset{\scriptstyle{\pm \textrm{0.28}}}{\textrm{82.40}}$ & $\underset{\scriptstyle{\pm \textrm{18.6}}}{\textrm{130.2}}$ &  & 0.93 & $\underset{\scriptstyle{\pm \textrm{0.29}}}{\textrm{84.13}}$ & $\underset{\scriptstyle{\pm \textrm{16.6}}}{\textrm{211.8}}$ &  & 0.96 & $\underset{\scriptstyle{\pm \textrm{0.27}}}{\textrm{83.51}}$ & $\underset{\scriptstyle{\pm \textrm{39.9}}}{\textrm{121.4}}$ &  & {3.8} & {3.0} \\
        & & \cellcolor{gray!20}\textbf{LLCG} & $M=3$ & & 0.88 & $\underset{\scriptstyle{\pm \textrm{0.31}}}{\textrm{47.87}}$ & $\underset{\scriptstyle{\pm \textrm{15.5}}}{\textrm{229.0}}$ &  & 0.95 & $\underset{\scriptstyle{\pm \textrm{0.02}}}{\textrm{81.88}}$ & $\underset{\scriptstyle{\pm \textrm{10.7}}}{\textrm{134.5}}$ &  & 0.93 & $\underset{\scriptstyle{\pm \textrm{0.10}}}{\textrm{84.43}}$ & $\underset{\scriptstyle{\pm \textrm{2.1}}}{\textrm{184.8}}$ &  & 0.96 & $\underset{\scriptstyle{\pm \textrm{0.40}}}{\textrm{83.14}}$ & $\underset{\scriptstyle{\pm \textrm{13.4}}}{\textrm{91.4}}$ &  & {3.5} & {3.5} \\
        & & \cellcolor{gray!20}\textbf{\methodmultigpu} & - & & 1.00 & $\underset{\scriptstyle{\pm \textrm{0.11}}}{\textrm{46.63}}$ & $\underset{\scriptstyle{\pm \textrm{4.3}}}{\textrm{47.5}}$ &  & 1.00 & $\underset{\scriptstyle{\pm \textrm{0.20}}}{\textrm{81.95}}$ & $\underset{\scriptstyle{\pm \textrm{0.8}}}{\textrm{173.4}}$ &  & 1.00 & $\underset{\scriptstyle{\pm \textrm{0.24}}}{\textrm{79.52}}$ & $\underset{\scriptstyle{\pm \textrm{0.0}}}{\textrm{240.0}}$ &  & 1.00 & $\underset{\scriptstyle{\pm \textrm{0.42}}}{\textrm{82.13}}$ & $\underset{\scriptstyle{\pm \textrm{0.3}}}{\textrm{87.4}}$ &  & {4.5} & {3.5} \\

        \bottomrule
        \end{tabular}
    }
    \vspace{-0.3cm}
\end{table*}

\begin{figure}[t]
    \centering
    \includegraphics[width=0.47\textwidth, trim=0.2cm 0 1.6cm 1.45cm, clip]{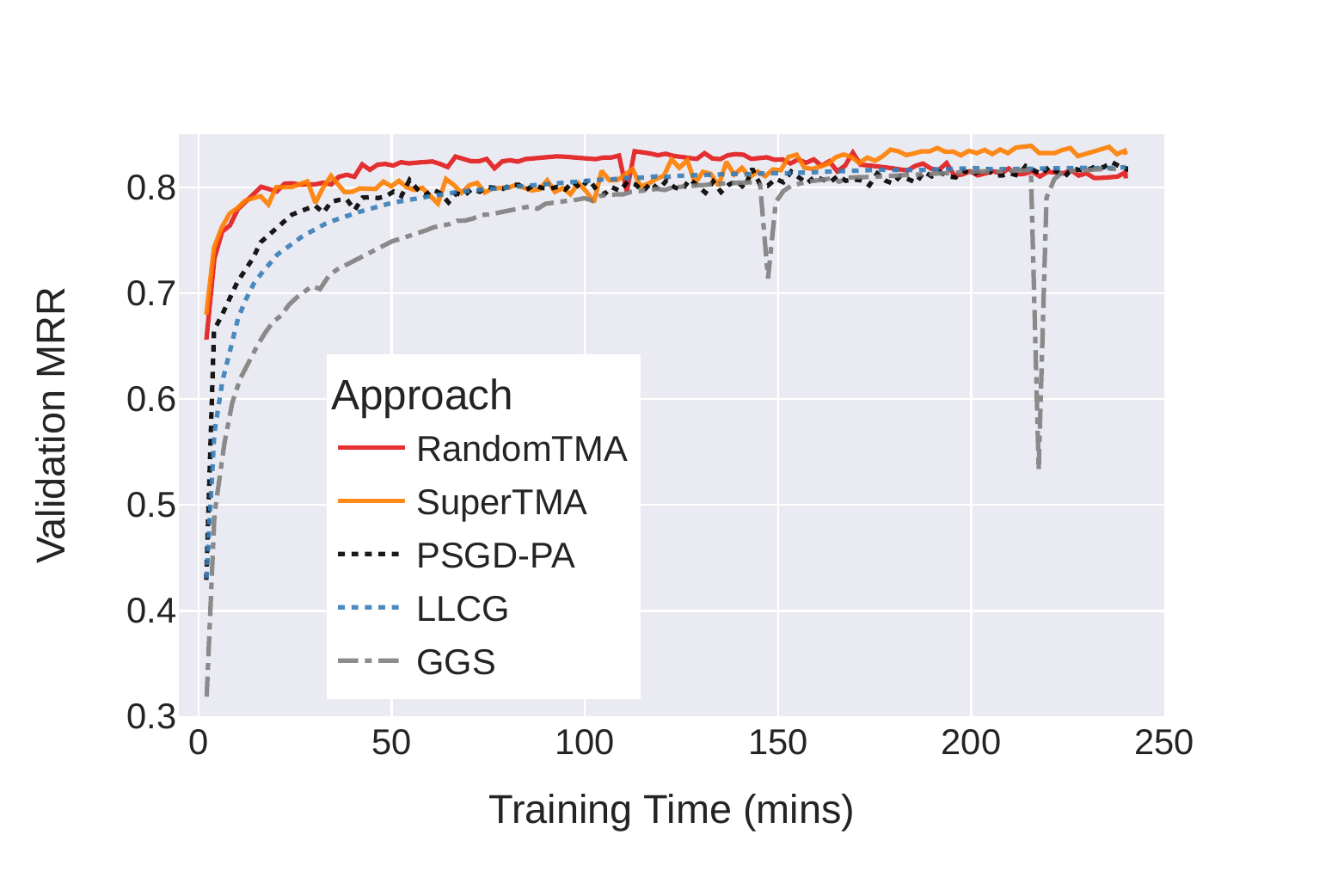}
    \vspace{-0.65cm}
    \caption{Validation MRR vs.\ training time for different training approaches on the best-performing GNN on ogbl-citation2. Table~\ref{tab:exp-benchmark} gives the test MRR and convergence time.
    }
    \label{fig:exp-benchmark-MRR}
    \vspace{-0.5cm}
\end{figure}

\begin{figure*}[t]
    \centering
    \includegraphics[width=\textwidth, trim=0 0 0 1.57cm, clip]{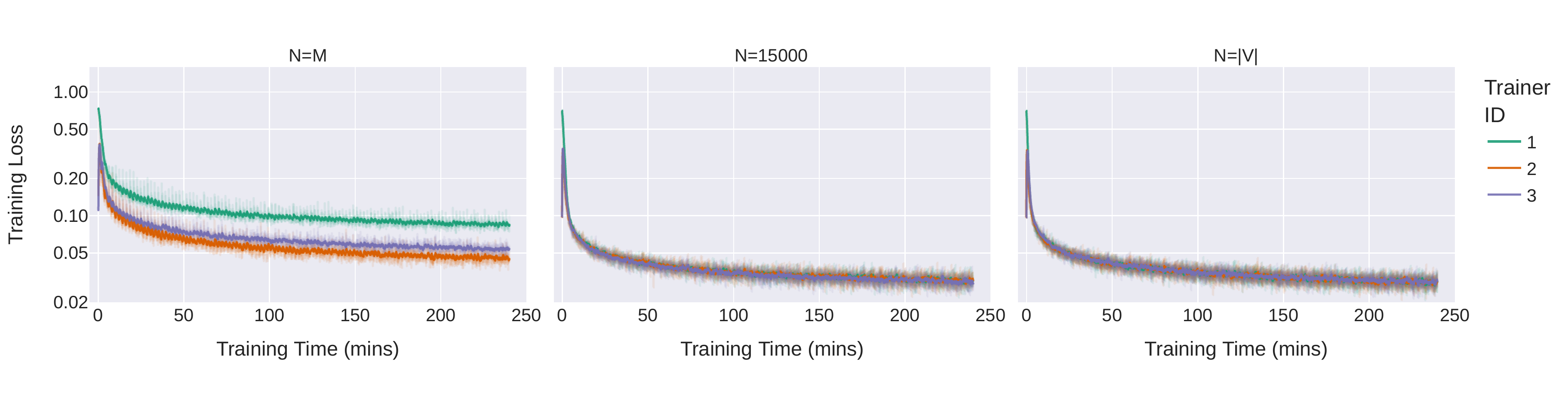}
    \captionsetup[subfigure]{skip=-0.7cm}
    \subcaptionbox{\methodbasic{} ($N=M$)\label{fig:exp-ablation-partition-loss-M}}{\hspace{0.31\linewidth}}
    \subcaptionbox{\method{} ($N=15000$)\label{fig:exp-ablation-partition-loss-15000}}{\hspace{0.31\linewidth}}
    \subcaptionbox{\methodrnd{} ($N=|\vertexSet|$)\label{fig:exp-ablation-partition-loss-V}}{\hspace{0.31\linewidth}}
    \vspace{-0.65cm}
    \caption{Training loss per trainer vs. training time for \methodbasic{}, \method{} and \methodrnd{} when training SAGE on MAG240M-P. We show the curves smoothed by exponential moving average ($\alpha = 0.1$), with raw curves dimmed in the background.
    }
    \label{fig:exp-ablation-partition-loss}
    \vspace{-0.15cm}
\end{figure*}

\paragraph{Setup} We compare the convergence speed and link prediction performance of the proposed \methodrnd{} and \method{} approaches with other baseline training approaches on different benchmark datasets. In Table~\ref{tab:exp-benchmark}, we list the best performance achieved by each approach with the best GNN encoder. {(Table~\ref{tab:exp-ablation-base-models} and \ref{tab:exp-ablation-base-models-ecomm} in the appendix provides the full results, and the graph partitioning runtime per approach, if applicable.)}
{For the convergence speed of each approach, we report the training time that each approach 
takes to reach within 1\% interval of its maximum validation MRR. 
We also list the ratio $r$ of the edges in the training graph that are available to each method.} We plot the change of validation MRR with respect to the training time on ogbl-citation2 in Fig.~\ref{fig:exp-benchmark-MRR}. 

\paragraph{Observations} 
We observe that despite having less training samples available due to increased cross-partition edges, 
\methodrnd{} and \method{} perform against the expectation of the previous work~\citep{ramezani2021learn} and achieve the best performance on each dataset and the highest average rankings in MRR. 
Moreover, the faster variant \methodrnd{} has the best convergence speed overall (it has the highest average ranking in convergence time) and is up to
2.31x faster than the fastest baseline, while still achieving comparable performance to \method{}. 
Overall, we find that \methodrnd{} strikes the best balance between performance and convergence speed, {while \method{} may be preferred in applications were the best possible task performance is critical}. The superior performance and convergence speed of \methodrnd{} and \method{} also demonstrate the effectiveness of our proposed partition schemes.

\subsection{(Q2) Advantages over Baselines}
To further dive into the reason behind the improved performance by our proposed approaches, we summarize and discuss two advantages of \method{} and \methodrnd{} over existing approaches.

\subsubsection{Reduced Discrepancy among Trainers with Randomized Partitions -- Empirical Validation of Theory.} We empirically validate our theoretical analysis by comparing the discrepancy of training losses among different trainers for \methodbasic{}, \method{} and \methodrnd{}, and show the plots in Fig.~\ref{fig:exp-ablation-partition-loss}. 
The usual $N=M$ partition scheme, which is adopted by baseline approaches like \methodbasic{}, leads to significant discrepancies among different trainers in the converged loss values, as shown in Fig.~\ref{fig:exp-ablation-partition-loss-M}, despite having the least cross-partition edges ignored in the training.
In comparison, both the super-node assignment ($N=15000$, Fig.~\ref{fig:exp-ablation-partition-loss-15000}) and random assignment ($N=|\vertexSet|$, Fig.~\ref{fig:exp-ablation-partition-loss-V}), adopted respectively by \method{} and \methodrnd{}, show better consistency of the converged loss values across multiple trainers; they also converge to smaller loss values compared to the classical \methodbasic{} partition scheme. 
We observe similar trends for other GNN models and datasets.
While smaller loss values on the training split do not always correspond to better performance on the validation and test splits, as the issue of overfitting can occur, the improved consistency of loss convergence across trainers explains the significant improvement in performance of the proposed approaches {over \methodbasic{} and \llcgbasic{}} and further validates Thm.~\ref{thm:homophily-grad-loss} on real-world datasets. 

\subsubsection{Improved Efficiency with TMA} 
\label{sec:exp-efficiency}
{We compare the efficiency of the proposed approaches to the baselines by measuring their GPU memory usage and the number 
of training steps finished on distributed trainers, and present the results on MAG240M-P dataset in Table~\ref{tab:exp-efficiency}.
These results demonstrate the improved efficiency of our Time-based Model Aggregation (TMA) mechanism:
all approaches with TMA (including \methodbasic{} and \llcgbasic{} that we enhanced) finish 2.69x to 6.45x more training steps on their \emph{slowest} trainer compared to \methodmultigpu{}, which conducts synchronous SGD after each training step.
Though the reduced size of the training graph on TMA approaches also contributes to reduced time per training step, the ratio of throughput improvement far exceeds the ratio of reduced graph size (which is reflected by the sampling ratio $r$ and GPU memory usage). 
Thus, we attribute %
the significantly improved efficiency of TMA-based approaches to the reduced overhead of synchronization among trainers enabled by the TMA mechanism: by eliminating the need for synchronization after each training step, TMA better accommodates the speed difference among trainers, which is up to 28.8\% as we show in Table~\ref{tab:exp-efficiency}. In comparison, the slowest trainer controls the training speed of the distributed system in \methodmultigpu{} (and also in the original design of step-based aggregation interval in \methodbasic{} and \llcgbasic{}), which results in significantly fewer completed training steps. 
We also note that our proposed approaches, \methodrnd{} and \method{}, by having the least GPU memory usage among all approaches due to the reduced  training graph size, enable better scalability to large datasets. 
}
{Overall, these results show the improved efficiency of our proposed approaches over the baselines (i.e., larger number of completed steps), which also contributes to improved performance and convergence speed.
}

\begin{table}[t]
    \centering
    \small
    \caption{Efficiency of training approaches: GPU memory usage, convergence time (min), and the range of the amount of training steps (in thousands) finished on distributed trainers. Results with the best efficiency are highlighted in green. As discussed in Sec.~\ref{sec:exp-setups}, we enhanced \methodbasic{} and \llcgbasic{} with our time-based model aggregation; \methodmultigpu{} uses synchronous SGD after each training step.}
    \label{tab:exp-efficiency}
    \vspace{-0.2cm}
    \resizebox{\columnwidth}{!}{
        \begin{tabular}{H c c H c H cc cHcr}
            \toprule
            
            \multirow{3.5}{*}{\rotatebox[origin=c]{90}{\textbf{\shortstack[c]{GNN}}}}
            & \multirow{2.5}{*}{\textbf{\shortstack[c]{Dataset\\\textnormal{($|\edgeSet|$,} \\ \textnormal{GNN)}}}}
            & \multirow{2.5}{*}{\textbf{\shortstack[c]{Train\\Approach}}} 
            & \multirow{3.5}{*}{\textbf{\shortstack[c]{\#Parts}}}
            & \multirow{2.5}{*}{\textbf{\shortstack[c]{Ratio\\\textnormal{($r$)}}}} &
            & \multirow{2.5}{*}{\textbf{\shortstack[c]{GPU\\RAM\\\textnormal{(GB)}}}}
            & \multirow{2.5}{*}{\textbf{\shortstack[c]{Conv.\\Time\\\textnormal{(min)}}}}
            & \multicolumn{4}{c}{\textbf{Step Finished ($10^3$)}} \\
            \cmidrule{9-12} 
            & & & & & & & 
            & \multicolumn{1}{c}{Min} & & {{Max}}  & \multicolumn{1}{c}{{Diff}} \\
            \midrule

            \multirow{4.5}{*}{\rotatebox[origin=c]{90}{\textbf{\shortstack[c]{SAGE}}}}
            & \multirow{7}{*}{\textbf{\shortstack[c]{MAG\\240M-P\\\textcolor{gray}{\textnormal{(1.30B,}}\\\textcolor{gray}{\textnormal{SAGE)}}}}}
            & \methodrnd & $|\vertexSet|$ & 0.33 & & \cellcolor{green!20}$\underset{\scriptstyle{\pm \textrm{0.03}}}{\textrm{7.98}}$ & \cellcolor{green!20}$\underset{\scriptstyle{\pm \textrm{27.6}}}{\textrm{169.3}}$ & \cellcolor{green!20}$\underset{\scriptstyle{\pm \textrm{0.39}}}{\textrm{6.64}}$ & $\underset{\scriptstyle{\pm \textrm{0.29}}}{\textrm{6.88}}$ & \cellcolor{green!20}$\underset{\scriptstyle{\pm \textrm{0.16}}}{\textrm{7.07}}$ & 6.1\% \\
            & & \method & 15,000 & 0.64 & & $\underset{\scriptstyle{\pm \textrm{0.01}}}{\textrm{9.32}}$ & $\underset{\scriptstyle{\pm \textrm{0.1}}}{\textrm{189.5}}$ & $\underset{\scriptstyle{\pm \textrm{0.05}}}{\textrm{4.74}}$ & $\underset{\scriptstyle{\pm \textrm{0.08}}}{\textrm{5.33}}$ & $\underset{\scriptstyle{\pm \textrm{0.16}}}{\textrm{5.70}}$ & 16.9\% \\
            \noalign{\vskip 0.25ex}
            \cdashline{3-12}[0.8pt/2pt]
            \noalign{\vskip 0.25ex}
            & & \cellcolor{gray!20}\methodbasic & $M$ & 0.93 & & $\underset{\scriptstyle{\pm \textrm{0.00}}}{\textrm{11.25}}$ & $\underset{\scriptstyle{\pm \textrm{16.6}}}{\textrm{211.8}}$ & $\underset{\scriptstyle{\pm \textrm{0.02}}}{\textrm{3.80}}$ & $\underset{\scriptstyle{\pm \textrm{0.16}}}{\textrm{4.61}}$ & $\underset{\scriptstyle{\pm \textrm{0.33}}}{\textrm{5.33}}$ & 28.8\% \\
            & & \cellcolor{gray!20}\llcgbasic & $M$ & 0.93 & & $\underset{\scriptstyle{\pm \textrm{0.04}}}{\textrm{11.30}}$ & $\underset{\scriptstyle{\pm \textrm{2.1}}}{\textrm{184.8}}$ & $\underset{\scriptstyle{\pm \textrm{0.01}}}{\textrm{4.08}}$ & $\underset{\scriptstyle{\pm \textrm{0.08}}}{\textrm{4.68}}$ & $\underset{\scriptstyle{\pm \textrm{0.01}}}{\textrm{5.32}}$ & 23.4\% \\
            \noalign{\vskip 0.25ex}
            \cdashline{3-12}[0.8pt/2pt]
            \noalign{\vskip 0.25ex}
            & & \cellcolor{gray!20}\methodmultigpu & - & 1.00 & & $\underset{\scriptstyle{\pm \textrm{0.01}}}{\textrm{12.12}}$ & $\underset{\scriptstyle{\pm \textrm{0.0}}}{\textrm{240.0}}$ & $\underset{\scriptstyle{\pm \textrm{0.07}}}{\textrm{1.03}}$ & $\underset{\scriptstyle{\pm \textrm{0.07}}}{\textrm{1.03}}$ & $\underset{\scriptstyle{\pm \textrm{0.07}}}{\textrm{1.03}}$ & 0.0\% \\

            \bottomrule
        \end{tabular}
   }
    \vspace{-0.4cm}
\end{table}

\subsection{(Q3) Robustness to Hyperparameters}

\subsubsection{Ablation on Aggregation Interval} 
For {approaches that leverage model aggregation (i.e., \methodrnd{}, \method{}, \methodbasic{} and \llcgbasic{})}, we examine the effect of the aggregation interval $\rho$ by varying it as 2  (default setting), 8 and 30 minutes. In Table~\ref{tab:exp-agg-interval}, we report the performance and convergence time under these scenarios, 
where we select the best-performing base model for each training approach and dataset {(cf.\ Table~\ref{tab:exp-ablation-base-models} for the best models)}.

\methodrnd{} and \method{} show consistent prediction performance regardless of the choice of the interval:
the differences in test MRR is less than 1\% and 0.2\% on Reddit and MAG240M-P, respectively. 
On the other hand, the baseline approaches \methodbasic{} and \llcgbasic{} show significant sensitivity to the interval:
as the aggregation interval increases, the test MRR drops for both methods by up to 7.88\% and 1.66\% on Reddit and MAG240M-P, respectively.
These observations show that \methodrnd{} and \method{}, thanks to our proposed partition schemes, do not require frequent aggregations like \methodbasic{} and \llcgbasic{} to achieve their peak performance and convergence speed, which enables further reduction of the communication overhead with longer intervals. 

\begin{table}[t]
    \centering
    \small
    \caption{Varying aggregation interval $\rho$: Comparison of link prediction performance ($\textrm{MRR}$) and convergence time (min). %
    Within each row, we highlight the aggregation interval with the best MRR in blue and the least convergence time in green.
    }
    \label{tab:exp-agg-interval}
    \vspace{-0.2cm}
    \resizebox{\columnwidth}{!}{
        \begin{tabular}{H c c H H H rrr c rrr rrr}
            \toprule
            
            \multirow{3.5}{*}{\rotatebox[origin=c]{90}{\textbf{\shortstack[c]{GNN}}}}
            & \multirow{2.5}{*}{\textbf{\shortstack[c]{Dataset\\\textnormal{($|\edgeSet|$,} \\ \textnormal{GNN)}}}}
            & \multirow{2.5}{*}{\textbf{\shortstack[c]{Train\\Approach}}} 
            & \multirow{3.5}{*}{\textbf{\shortstack[c]{\#Parts}}}
            & \multirow{3.5}{*}{\textbf{\shortstack[c]{Ratio\\\textnormal{($r$)}}}} &
            \multicolumn{4}{c}{\textbf{Test MRR (\%)}} &  
            & \multicolumn{3}{c}{\textbf{Conv. Time} (min)}
            \\
             \cmidrule{6-9} \cmidrule{11-13} 
            & & & & &
                    & \multicolumn{1}{c}{$\rho$=2} & \multicolumn{1}{c}{{$\rho$=8}} & \multicolumn{1}{c}{{$\rho$=30}} &
                    & \multicolumn{1}{c}{{$\rho$=2}} & \multicolumn{1}{c}{{$\rho$=8}} & \multicolumn{1}{c}{{$\rho$=30}} \\
            \midrule

            \multirow{4.5}{*}{\rotatebox[origin=c]{90}{\textbf{\shortstack[c]{GCN}}}}
            & \multirow{6}{*}{\textbf{\shortstack[c]{Reddit\\\textcolor{gray}{\textnormal{(114M,}}\\\textcolor{gray}{\textnormal{GCN)}}}}}
            & \methodrnd & $|\vertexSet|$ & 0.33 & & $\underset{\scriptstyle{\pm \textrm{0.21}}}{\textrm{47.78}}$\cellcolor{blue!30} & $\underset{\scriptstyle{\pm \textrm{0.60}}}{\textrm{47.38}}$ & $\underset{\scriptstyle{\pm \textrm{0.47}}}{\textrm{46.86}}$ &  & $\underset{\scriptstyle{\pm \textrm{7.1}}}{\textrm{67.4}}$ & $\underset{\scriptstyle{\pm \textrm{11.3}}}{\textrm{40.1}}$\cellcolor{green!30} & $\underset{\scriptstyle{\pm \textrm{0.0}}}{\textrm{60.0}}$ \\
            & & \method & 15,000 & 0.35 & & $\underset{\scriptstyle{\pm \textrm{0.64}}}{\textrm{48.68}}$\cellcolor{blue!30} & $\underset{\scriptstyle{\pm \textrm{0.09}}}{\textrm{48.51}}$ & $\underset{\scriptstyle{\pm \textrm{0.09}}}{\textrm{47.77}}$ &  & $\underset{\scriptstyle{\pm \textrm{6.9}}}{\textrm{154.4}}$ & $\underset{\scriptstyle{\pm \textrm{51.0}}}{\textrm{76.1}}$ & $\underset{\scriptstyle{\pm \textrm{21.2}}}{\textrm{75.0}}$\cellcolor{green!30} \\
            \noalign{\vskip 0.25ex}
            \cdashline{3-14}[0.8pt/2pt]
            \noalign{\vskip 0.25ex}
            & & \cellcolor{gray!20}\methodbasicshort & $M$ & 0.88 & & $\underset{\scriptstyle{\pm \textrm{0.35}}}{\textrm{46.02}}$\cellcolor{blue!30} & $\underset{\scriptstyle{\pm \textrm{0.34}}}{\textrm{43.78}}$ & $\underset{\scriptstyle{\pm \textrm{0.13}}}{\textrm{40.21}}$ &  & $\underset{\scriptstyle{\pm \textrm{10.0}}}{\textrm{37.2}}$\cellcolor{green!30} & $\underset{\scriptstyle{\pm \textrm{51.0}}}{\textrm{188.3}}$ & $\underset{\scriptstyle{\pm \textrm{63.7}}}{\textrm{165.1}}$ \\
            & & \cellcolor{gray!20}\llcgbasic & $M$ & 0.88 & & $\underset{\scriptstyle{\pm \textrm{0.31}}}{\textrm{47.87}}$\cellcolor{blue!30} & $\underset{\scriptstyle{\pm \textrm{0.19}}}{\textrm{44.54}}$ & $\underset{\scriptstyle{\pm \textrm{0.61}}}{\textrm{39.99}}$ &  & $\underset{\scriptstyle{\pm \textrm{15.5}}}{\textrm{229.0}}$ & $\underset{\scriptstyle{\pm \textrm{11.3}}}{\textrm{48.2}}$\cellcolor{green!30} & $\underset{\scriptstyle{\pm \textrm{63.7}}}{\textrm{165.2}}$ \\
            
            \midrule
            
            \multirow{4.5}{*}{\rotatebox[origin=c]{90}{\textbf{\shortstack[c]{SAGE}}}}
            & \multirow{6}{*}{\textbf{\shortstack[c]{MAG\\240M-P\\\textcolor{gray}{\textnormal{(1.30B,}}\\\textcolor{gray}{\textnormal{SAGE)}}}}}
            & \methodrnd & $|\vertexSet|$ & 0.33 & & $\underset{\scriptstyle{\pm \textrm{0.09}}}{\textrm{85.77}}$ & $\underset{\scriptstyle{\pm \textrm{0.17}}}{\textrm{85.79}}$ & $\underset{\scriptstyle{\pm \textrm{0.26}}}{\textrm{85.82}}$\cellcolor{blue!30} &  & $\underset{\scriptstyle{\pm \textrm{27.6}}}{\textrm{169.3}}$ & $\underset{\scriptstyle{\pm \textrm{5.6}}}{\textrm{164.9}}$\cellcolor{green!30} & $\underset{\scriptstyle{\pm \textrm{0.0}}}{\textrm{180.2}}$ \\
            & & \method & 15,000 & 0.64 & & $\underset{\scriptstyle{\pm \textrm{0.36}}}{\textrm{85.27}}$ & $\underset{\scriptstyle{\pm \textrm{0.25}}}{\textrm{85.38}}$\cellcolor{blue!30} & $\underset{\scriptstyle{\pm \textrm{0.10}}}{\textrm{85.22}}$ &  & $\underset{\scriptstyle{\pm \textrm{0.1}}}{\textrm{189.5}}$\cellcolor{green!30} & $\underset{\scriptstyle{\pm \textrm{11.4}}}{\textrm{193.2}}$ & $\underset{\scriptstyle{\pm \textrm{0.0}}}{\textrm{210.3}}$ \\
            \noalign{\vskip 0.25ex}
            \cdashline{3-14}[0.8pt/2pt]
            \noalign{\vskip 0.25ex}
            & & \cellcolor{gray!20}\methodbasicshort & $M$ & 0.93 & & $\underset{\scriptstyle{\pm \textrm{0.29}}}{\textrm{84.13}}$\cellcolor{blue!30} & $\underset{\scriptstyle{\pm \textrm{0.25}}}{\textrm{83.44}}$ & $\underset{\scriptstyle{\pm \textrm{0.39}}}{\textrm{82.47}}$ &  & $\underset{\scriptstyle{\pm \textrm{16.6}}}{\textrm{211.8}}$ & $\underset{\scriptstyle{\pm \textrm{45.7}}}{\textrm{201.4}}$\cellcolor{green!30} & $\underset{\scriptstyle{\pm \textrm{0.0}}}{\textrm{210.3}}$ \\
            & & \cellcolor{gray!20}\llcgbasic & $M$ & 0.93 & & $\underset{\scriptstyle{\pm \textrm{0.10}}}{\textrm{84.43}}$\cellcolor{blue!30} & $\underset{\scriptstyle{\pm \textrm{0.40}}}{\textrm{84.27}}$ & $\underset{\scriptstyle{\pm \textrm{0.12}}}{\textrm{82.95}}$ &  & $\underset{\scriptstyle{\pm \textrm{2.1}}}{\textrm{184.8}}$\cellcolor{green!30} & $\underset{\scriptstyle{\pm \textrm{6.0}}}{\textrm{191.9}}$ & $\underset{\scriptstyle{\pm \textrm{0.2}}}{\textrm{211.3}}$ \\

            \bottomrule
        \end{tabular}
   }
    \vspace{-0.5cm}
\end{table}

\subsubsection{Ablation on Number of Trainers} 
\label{sec:exp-ablation-num-trainers}

\begin{table*}[t]
    \centering
    \small
    \caption{Varying number of trainers $M$: Comparison of ratio $r$ of training samples available, link prediction performance ($\textrm{MRR}$), and convergence time (min).
    Within each row, we highlight 
    the number of trainers with 
    the best MRR in blue and the least convergence time in green. 
    ``OOM'' denotes that experiments run out of memory.%
    }
    \label{tab:exp-num-trainers}
    \vspace{-0.3cm}
        \begin{tabular}{H c c H H ccc c ccc c ccc HHH}
            \toprule
            
             \multirow{3.5}{*}{\rotatebox[origin=c]{90}{\textbf{\shortstack[c]{GNN}}}}
            & \multirow{2.5}{*}{\textbf{\shortstack[c]{Dataset\\\textnormal{($|\edgeSet|$,} \\ \textnormal{GNN)}}}}
            & \multirow{2.5}{*}{\textbf{\shortstack[c]{Train\\Approach}}} 
            & \multirow{3.5}{*}{\textbf{\shortstack[c]{\#Parts}}}
            & \multirow{3.5}{*}{\textbf{\shortstack[c]{Ratio\\\textnormal{($r$)}}}} 
            & \multicolumn{3}{c}{\textbf{Ratio} ($r$)} & & \multicolumn{3}{c}{\textbf{Test MRR} (\%)} & & \multicolumn{3}{c}{\textbf{Conv. Time} (min)}\\
             \cmidrule{6-8} \cmidrule{10-12} \cmidrule{14-16} 
            & & & & & $M$=3 & \multicolumn{1}{c}{$M$=5} & \multicolumn{1}{c}{$M$=23} &
                  & \multicolumn{1}{c}{$M$=3} & \multicolumn{1}{c}{$M$=5} & \multicolumn{1}{c}{$M$=23} &
                  & \multicolumn{1}{c}{$M$=3} & \multicolumn{1}{c}{$M$=5} & \multicolumn{1}{c}{$M$=23} \\
            \midrule

            \multirow{4}{*}{\rotatebox[origin=c]{90}{\textbf{\shortstack[c]{SAGE}}}}
            & \multirow{4}{*}{\textbf{\shortstack[c]{MAG\\240M-P\\\textcolor{gray}{\textnormal{(1.30B,}}\\\textcolor{gray}{\textnormal{SAGE)}}}}}
            & \methodrnd & $|\vertexSet|$ & 0.33 & 0.33 & 0.20 & 0.04 &  & 85.77\tiny{$\pm$0.09} & 85.97\tiny{$\pm$0.32}\cellcolor{blue!30} & 84.94\tiny{$\pm$0.23} &  & 169.3\tiny{$\pm$27.6} & 158.4\tiny{$\pm$4.1} & 125.0\tiny{$\pm$5.2}\cellcolor{green!30} \\
            & & \method & 15,000 & 0.64 & 0.64 & 0.56 & 0.48 &  & 85.27\tiny{$\pm$0.36} & 86.02\tiny{$\pm$0.27} & 86.21\tiny{$\pm$0.53}\cellcolor{blue!30} &  & 189.5\tiny{$\pm$0.1} & 181.8\tiny{$\pm$9.8}\cellcolor{green!30} & 190.6\tiny{$\pm$16.3} \\
            \noalign{\vskip 0.25ex}
            \cdashline{3-16}[0.8pt/2pt]
            \noalign{\vskip 0.25ex}
            & & \cellcolor{gray!20}\methodbasicshort & $M$ & 0.93 & 0.93 & 0.90 & 0.78 &  & 84.13\tiny{$\pm$0.29} & 84.35\tiny{$\pm$0.07}\cellcolor{blue!30} & 82.13\tiny{$\pm$0.12} &  & 211.8\tiny{$\pm$16.6} & 206.5\tiny{$\pm$0.1}\cellcolor{green!30} & 208.8\tiny{$\pm$17.0} \\
            & & \cellcolor{gray!20}\llcgbasic & $M$ & 0.93 & 0.93 & 0.90 & 0.90 &  & 84.43\tiny{$\pm$0.10}\cellcolor{blue!30} & 83.87\tiny{$\pm$0.39} & (OOM) &  & 184.8\tiny{$\pm$2.1}\cellcolor{green!30} & 194.6\tiny{$\pm$16.6} & (OOM) \\
            
            \midrule
            
            \multirow{4}{*}{\rotatebox[origin=c]{90}{\textbf{\shortstack[c]{SAGE}}}}
            & \multirow{4}{*}{\textbf{\shortstack[c]{E-comm\\\textcolor{gray}{\textnormal{(207M,}}\\\textcolor{gray}{\textnormal{GCN)}}}}}
            & \methodrnd & $|\vertexSet|$ & r & 0.33 & 0.20 & 0.04 &  & 84.12\tiny{$\pm$0.02} & 84.95\tiny{$\pm$0.41}\cellcolor{blue!30} & 80.73\tiny{$\pm$0.06} &  & 52.5\tiny{$\pm$20.0} & 67.0\tiny{$\pm$23.0} & 18.4\tiny{$\pm$14.5}\cellcolor{green!30} \\
            & & \method & 15,000 & r & 0.76 & 0.71 & 0.65 &  & 84.44\tiny{$\pm$0.45} & 84.95\tiny{$\pm$0.27} & 85.48\tiny{$\pm$0.37}\cellcolor{blue!30} &  & 126.3\tiny{$\pm$12.8}\cellcolor{green!30} & 129.0\tiny{$\pm$19.1} & 130.5\tiny{$\pm$2.3} \\
            \noalign{\vskip 0.25ex}
            \cdashline{3-16}[0.8pt/2pt]
            \noalign{\vskip 0.25ex}
            & & \cellcolor{gray!20}\methodbasicshort & $M$ & r & 0.96 & 0.96 & 0.92 &  & 83.51\tiny{$\pm$0.27} & 83.55\tiny{$\pm$0.29}\cellcolor{blue!30} & 83.40\tiny{$\pm$0.01} &  & 121.4\tiny{$\pm$39.9} & 124.0\tiny{$\pm$11.7} & 107.6\tiny{$\pm$0.1}\cellcolor{green!30} \\
            & & \cellcolor{gray!20}\llcgbasic & $M$ & r & 0.96 & 0.96 & 0.92 &  & 83.14\tiny{$\pm$0.40} & 83.46\tiny{$\pm$0.37} & 83.93\tiny{$\pm$0.55}\cellcolor{blue!30} &  & 91.4\tiny{$\pm$13.4}\cellcolor{green!30} & 124.1\tiny{$\pm$32.2} & 111.1\tiny{$\pm$10.2} \\

            \bottomrule
        \end{tabular}
    \vspace{-0.2cm}
\end{table*}

To understand the effect of increased number of trainers for model aggregation approaches, we compare the performance, convergence time and ratio of training samples available %
in the cases of $M=3$ (the default setting), $M=5$, {and a very large number of $M=23$ trainers\footnote{{Number of trainers $M=23$ maps to the maximum number of trainers we can set up with 24 GPUs, as we reserve one GPU for model evaluation on the server process.}}} in Table~\ref{tab:exp-num-trainers}.
We run this experiment on the largest MAG240M-P dataset and {the proprietary large E-comm dataset}, and select the best-performing GNN model {(i.e., GraphSAGE for MAG240M-P, and GCN with MLP decoder for E-comm)} for all training approaches as the base model. 

{We observe that the amount of available training samples decreases for all approaches as the number of trainers increases, due to the increase of cross-partition edges. 
\methodrnd{} has a sweet spot for edge ratio $r$ and the number of trainers $M$: compared to $M=3$, it shows slightly improved performance for $M=5$, but worse performance for $M=23$, especially for the smaller E-comm dataset. We attribute this to the trade-off between increased data throughput and decreased amount of training samples for an increased number of trainers.}
{\method{}, on the other hand, effectively mitigates the side-effect of data loss under increased number of trainers, has significantly more training samples compared to \methodrnd{}, and shows consistently improved performance; this demonstrates the effectiveness of conducting randomized partitions on mini-clusters. %
Despite leveraging the most training edges under all cases, \methodbasic{} and \llcgbasic{} consistently perform worse than \method{} (and \methodrnd{} in most cases), which highlights the importance of data uniformity over the amount of available training samples.}

\vspace{-0.2cm}
\subsection{(Q4) Robustness to Trainer Failures}
\label{sec:exp-failure}

Distributed systems can suffer from failure of workers as a result of unexpected faults or issues with the communication network. 
Fortunately, model aggregation training allows the frameworks to be robust to partial failures (e.g., when  some trainers go offline), as the training can continue with only the remaining trainers. 
However, the subgraphs assigned to failed instances will be unavailable in the remaining of the training process, unless the server reassigns these subgraphs to any available back-up training instances. 

\paragraph{Setup} Here we emulate a simple scenario of failure where $F=1$ of the $M=3$ trainers in previous experiments fail to start, with no back-up trainer in place;
in this case, we complete the training with the remaining graph information on $M-1$ trainers. 
Our goal is to understand the robustness of \methodrnd{} and \method{} to trainer failures in comparison with  model aggregation baselines. 

Table~\ref{tab:exp-failures} reports the performance and convergence time of the training approaches when a worker fails to start ($F = 1$), compared with the case where all workers proceed normally ($F = 0$). For the $F = 1$ case, we run the $M$ experiments per random seed by dropping a different partition at a time to emulate failure of different trainers under the same assignment, and report the average results. 

\paragraph{Observations} We observe that the performance and convergence speed of \methodrnd{} and \method{} are more robust to the failure of the trainers compared to \methodbasic{} and \llcgbasic{}: the test MRR  decreases by less than 0.3\% for \methodrnd{} and \method{} as a result of the single trainer failure; in comparison, the test MRR of \methodbasic{} and \llcgbasic{} decreases more than 2.0\% in the case of failure. 
These results demonstrate the improved robustness of \methodrnd{} and \method{}: less discrepancy among data assigned to different trainers minimizes the information loss in the case of failures.

\begin{table}[t]
    \centering
    \small
    \caption{Robustness to trainer failures: Comparison of link prediction performance ($\textrm{MRR}$) and convergence time (min) when one of the $M=3$ trainers fails to start. 
    For $F = 1$, we run $M$ experiments per random seed by dropping a different subgraph at a time, and report the average metrics. %
    }
    \label{tab:exp-failures}
    \vspace{-0.4cm}
        \begin{tabular}{H c c H H c>{\color{dark-gray}}c c c>{\color{dark-gray}}c H HH HHH}
            \toprule
            
            \multirow{3.5}{*}{\textbf{GNN}}
            & \multirow{2.5}{*}{\textbf{\shortstack[c]{Dataset\\\textnormal{($|\edgeSet|$,} \\ \textnormal{GNN)}}}}
            & \multirow{2.5}{*}{\textbf{\shortstack[c]{Train\\Approach}}} 
            & \multirow{3.5}{*}{\textbf{\shortstack[c]{\#Parts}}}
            & \multirow{3.5}{*}{\textbf{\shortstack[c]{Ratio\\\textnormal{($r$)}}}} & 
            \multicolumn{2}{c}{\textbf{Test MRR} (\%)} & & \multicolumn{2}{c}{\textbf{Conv. Time} (min)}\\
             \cmidrule{6-7} \cmidrule{9-10} \cmidrule{12-13} \cmidrule{15-16}
            & & & & & $F=1$ & \multicolumn{1}{c}{$F=0$} & 
                  & \multicolumn{1}{c}{$F=1$} & \multicolumn{1}{c}{$F=0$} \\
            \midrule
            
            \multirow{6}{*}{{\textbf{\shortstack[c]{SAGE}}}}
            & \multirow{5.5}{*}{\textbf{\shortstack[c]{MAG\\240M-P\\\textcolor{gray}{\textnormal{(1.30B,}}\\\textcolor{gray}{\textnormal{SAGE)}}}}}
            & \methodrnd & $|\vertexSet|$ & 0.33 & $\underset{\scriptstyle{\pm \textrm{0.08}}}{\textrm{85.54}}$ & $\underset{\scriptstyle{\pm \textrm{0.09}}}{\textrm{85.77}}$ &  & $\underset{\scriptstyle{\pm \textrm{13.6}}}{\textrm{161.8}}$ & $\underset{\scriptstyle{\pm \textrm{27.6}}}{\textrm{169.3}}$ \\
            & & \method & 15,000 & 0.64 & $\underset{\scriptstyle{\pm \textrm{0.11}}}{\textrm{85.17}}$ & $\underset{\scriptstyle{\pm \textrm{0.36}}}{\textrm{85.27}}$ &  & $\underset{\scriptstyle{\pm \textrm{12.5}}}{\textrm{191.7}}$ & $\underset{\scriptstyle{\pm \textrm{0.1}}}{\textrm{189.5}}$ \\
            \noalign{\vskip 0.25ex}
            \cdashline{3-14}[0.8pt/2pt]
            \noalign{\vskip 0.25ex}
            & & \cellcolor{gray!20}\methodbasic & $M$ & 0.93 & $\underset{\scriptstyle{\pm \textrm{4.09}}}{\textrm{82.09}}$ & $\underset{\scriptstyle{\pm \textrm{0.29}}}{\textrm{84.13}}$ &  & $\underset{\scriptstyle{\pm \textrm{16.0}}}{\textrm{199.0}}$ & $\underset{\scriptstyle{\pm \textrm{16.6}}}{\textrm{211.8}}$ \\
            & & \cellcolor{gray!20}\llcgbasic{} & - & 1.00 & $\underset{\scriptstyle{\pm \textrm{3.45}}}{\textrm{82.20}}$ & $\underset{\scriptstyle{\pm \textrm{0.10}}}{\textrm{84.43}}$ &  & $\underset{\scriptstyle{\pm \textrm{22.3}}}{\textrm{203.2}}$ & $\underset{\scriptstyle{\pm \textrm{2.1}}}{\textrm{184.8}}$ \\

            \bottomrule
        \end{tabular}
    \vspace{-0.4cm}
\end{table}

\vspace{-0.0cm}
\section{Conclusion}
\label{sec:conc}

In this work, we revisited prior assumptions that relate the performance of distributed GNN training under data parallelism with the coverage of cross-instance node dependencies, and surprisingly discovered that ignoring more cross-instance edges does not necessarily lead to decreased GNN performance. 
We theoretically analyzed the reason behind this phenomenon, and showed that the discrepancy of data distributions among different partitions caused by min-cut partitioning algorithms is more critical for the performance %
than the number of ignored cross-instance edges. 
Based on this finding, we proposed two randomized partition schemes on nodes and super-nodes that minimize the data discrepancy among instances; we also combined them with a simplified distributed GNN training framework that 
allows only local data access per trainer, and synchronizes the learned local models by conducting periodic, time-based model aggregation (TMA) to accommodate imbalanced loads and training speeds among trainers. %
We conducted extensive link prediction experiments on 
large-scale social, collaboration and e-commerce networks with up to 1.3 billion edges, and demonstrated that---despite ignoring more cross-instance dependencies---our proposed 
approaches, \methodrnd{} and \method{}, achieve state-of-the-art performance, are up to 2.31x faster at converging than the most efficient baseline, and show better robustness to trainer failures. 
{Future directions include evaluating our framework on other graph learning tasks such as node classification.}

\balance
{\small
\bibliographystyle{ACM-Reference-Format}
\bibliography{BIB/abbrev,BIB/all,BIB/all_career,BIB/main}
}

\newpage

\appendix

\twocolumn[\section*{\LARGE \centering Appendix}
\vspace{0.4cm}
]
\begingroup
\begin{table*}[t]
    \centering
    \caption{Comparison of link prediction performance ($\textrm{MRR}$) and convergence time (minutes) under different base models (i.e., GCN, GraphSAGE and MLP).
    The {convergence time is reported as the time to reach within 1\% interval of the maximum validation MRR.}
    We also report for each dataset and approach the ratio $r$ of the edges in the training graph that are available, and the preprocessing time of METIS (if needed). 
    We highlight within each row the GNN model (excluding MLP) with the best MRR in blue, and the least convergence time in green. 
    MLP is graph-agnostic and thus not tested for \llcgbasic{}. 
    }
    \label{tab:exp-ablation-base-models}
    {\small 
    \vspace{-0.3cm}
        \begin{tabular}{c c c c c cc>{\color{dark-gray}}c c rr>{\color{dark-gray}}r rrr}
        \toprule
        
        \multirow{2.5}{*}{\textbf{\shortstack[c]{Dataset\\\textnormal{($|\edgeSet|$)}}}}
        & \multirow{2.5}{*}{\textbf{\shortstack[c]{Train\\Approach}}} 
        & \multirow{2.5}{*}{\textbf{\shortstack[c]{\#Parts\\($N$)}}}
        & \multirow{2.5}{*}{\textbf{\shortstack[c]{Ratio \textnormal{($r$)}}}} 
        & \multirow{2.5}{*}{\textbf{\shortstack[c]{Prep.\\Time\\\textnormal{\small{(mins)}}}}}
        & \multicolumn{3}{c}{\textbf{Test MRR} (\%)} &  
        & \multicolumn{3}{c}{\textbf{Conv. Time} (min)}
        \\
         \cmidrule{6-8} \cmidrule{10-12} 
        & & & & 
                & \multicolumn{1}{c}{\textbf{GCN}} & \multicolumn{1}{c}{\textbf{SAGE}} & \multicolumn{1}{c}{\textbf{MLP}} &
                & \multicolumn{1}{c}{\textbf{GCN}} & \multicolumn{1}{c}{\textbf{SAGE}} & \multicolumn{1}{c}{\textbf{MLP}} \\
        \midrule

        \multirow{5}{*}{\textbf{\shortstack[c]{Reddit\\\textcolor{gray}{\textnormal{(114M)}}}}} & \textbf{\methodrnd} & $|\vertexSet|$ & 0.33 & 0 & 47.78\footnotesize{$\pm$0.21}\cellcolor{blue!20} & 43.65\footnotesize{$\pm$0.49} & 22.92\footnotesize{$\pm$0.02} &  & 67.4\footnotesize{$\pm$7.1}\cellcolor{green!20} & 175.9\footnotesize{$\pm$22.7} & 166.9\footnotesize{$\pm$31.2} \\
        & \textbf{\method} & 15,000 & 0.35 & 6.5 & 48.68\footnotesize{$\pm$0.64}\cellcolor{blue!20} & 43.93\footnotesize{$\pm$0.03} & 22.89\footnotesize{$\pm$0.15} &  & 154.4\footnotesize{$\pm$6.9}\cellcolor{green!20} & 188.9\footnotesize{$\pm$4.2} & 194.1\footnotesize{$\pm$9.9} \\
        \noalign{\vskip 0.25ex}
        \cdashline{2-14}[0.8pt/2pt]
        \noalign{\vskip 0.25ex}
        & \cellcolor{gray!20}\textbf{\methodbasic} & \cellcolor{gray!20}$M=3$ & \cellcolor{gray!20}0.88 & 0.7 & 46.02\footnotesize{$\pm$0.35}\cellcolor{blue!20} & 43.42\footnotesize{$\pm$2.08} & 16.55\footnotesize{$\pm$0.38} &  & 37.2\footnotesize{$\pm$10.0}\cellcolor{green!20} & 240.0\footnotesize{$\pm$0.0} & 240.0\footnotesize{$\pm$0.0} \\
        & \cellcolor{gray!20}\textbf{LLCG} & \cellcolor{gray!20}$M=3$ & \cellcolor{gray!20}0.88 & 0.7 & 47.87\footnotesize{$\pm$0.31}\cellcolor{blue!20} & 44.61\footnotesize{$\pm$0.14} & \multicolumn{1}{c}{-} &  & 229.0\footnotesize{$\pm$15.5} & 185.3\footnotesize{$\pm$77.4}\cellcolor{green!20} & \multicolumn{1}{c}{-} \\
        & \cellcolor{gray!20}\textbf{\methodmultigpu} & \cellcolor{gray!20}- & \cellcolor{gray!20}1.00 & 0 & 46.63\footnotesize{$\pm$0.11}\cellcolor{blue!20} & 43.85\footnotesize{$\pm$0.22} & 24.31\footnotesize{$\pm$0.09} &  & 47.5\footnotesize{$\pm$4.3}\cellcolor{green!20} & 209.5\footnotesize{$\pm$18.0} & 129.5\footnotesize{$\pm$14.6} \\
        
        \midrule
        
        \multirow{5}{*}{\textbf{\shortstack[c]{ogbl-\\citation2\\\textcolor{gray}{\textnormal{(30.5M)}}}}} & \textbf{\methodrnd} & $|\vertexSet|$ & 0.33 & 0 & 83.28\footnotesize{$\pm$0.24}\cellcolor{blue!20} & 80.96\footnotesize{$\pm$0.00} & 40.69\footnotesize{$\pm$0.01} &  & 56.4\footnotesize{$\pm$14.3}\cellcolor{green!20} & 101.9\footnotesize{$\pm$12.9} & 57.3\footnotesize{$\pm$12.8} \\
        & \textbf{\method} & 15,000 & 0.58 & 1.6 & 83.75\footnotesize{$\pm$0.43}\cellcolor{blue!20} & 80.90\footnotesize{$\pm$0.01} & 41.36\footnotesize{$\pm$0.08} &  & 126.8\footnotesize{$\pm$39.6} & 100.2\footnotesize{$\pm$18.6}\cellcolor{green!20} & 86.4\footnotesize{$\pm$11.4} \\
        \noalign{\vskip 0.25ex}
        \cdashline{2-14}[0.8pt/2pt]
        \noalign{\vskip 0.25ex}
        & \cellcolor{gray!20}\textbf{\methodbasic} & \cellcolor{gray!20}$M=3$ & \cellcolor{gray!20}0.95 & 0.7 & 82.40\footnotesize{$\pm$0.28}\cellcolor{blue!20} & 81.64\footnotesize{$\pm$0.00} & 39.43\footnotesize{$\pm$0.16} &  & 130.2\footnotesize{$\pm$18.6}\cellcolor{green!20} & 146.0\footnotesize{$\pm$8.8} & 176.1\footnotesize{$\pm$24.3} \\
        & \cellcolor{gray!20}\textbf{LLCG} & \cellcolor{gray!20}$M=3$ & \cellcolor{gray!20}0.95 & 0.7 & 81.62\footnotesize{$\pm$0.47} & 81.88\footnotesize{$\pm$0.02}\cellcolor{blue!20} & \multicolumn{1}{c}{-} &  & 142.6\footnotesize{$\pm$48.2} & 134.5\footnotesize{$\pm$10.7}\cellcolor{green!20} & \multicolumn{1}{c}{-} \\
        & \cellcolor{gray!20}\textbf{\methodmultigpu} & \cellcolor{gray!20}- & \cellcolor{gray!20}1.00 & 0 & 81.64\footnotesize{$\pm$0.17} & 81.95\footnotesize{$\pm$0.20}\cellcolor{blue!20} & 41.71\footnotesize{$\pm$0.03} &  & 178.5\footnotesize{$\pm$43.8} & 173.4\footnotesize{$\pm$0.8}\cellcolor{green!20} & 90.1\footnotesize{$\pm$9.9} \\
        
        \midrule
        
        \multirow{5}{*}{\textbf{\shortstack[c]{MAG\\240M-\\Papers\\\textcolor{gray}{\textnormal{(1.30B)}}}}} & \textbf{\methodrnd} & $|\vertexSet|$ & 0.33 & 0 & 85.08\footnotesize{$\pm$0.30} & 85.77\footnotesize{$\pm$0.09}\cellcolor{blue!20} & 48.54\footnotesize{$\pm$0.27} &  & 213.8\footnotesize{$\pm$23.2} & 169.3\footnotesize{$\pm$27.6}\cellcolor{green!20} & 156.1\footnotesize{$\pm$10.9} \\
        & \textbf{\method} & 15,000 & 0.64 & 153.9 & 82.21\footnotesize{$\pm$0.04} & 85.27\footnotesize{$\pm$0.36}\cellcolor{blue!20} & 49.14\footnotesize{$\pm$0.15} &  & 214.4\footnotesize{$\pm$36.2} & 189.5\footnotesize{$\pm$0.1}\cellcolor{green!20} & 164.8\footnotesize{$\pm$5.7} \\
        \noalign{\vskip 0.25ex}
        \cdashline{2-14}[0.8pt/2pt]
        \noalign{\vskip 0.25ex}
        & \cellcolor{gray!20}\textbf{\methodbasic} & \cellcolor{gray!20}$M=3$ & \cellcolor{gray!20}0.93 & 84.4 & 80.90\footnotesize{$\pm$0.09} & 84.13\footnotesize{$\pm$0.29}\cellcolor{blue!20} & 48.30\footnotesize{$\pm$0.15} &  & 240.0\footnotesize{$\pm$0.0} & 211.8\footnotesize{$\pm$16.6}\cellcolor{green!20} & 195.7\footnotesize{$\pm$17.6} \\
        & \cellcolor{gray!20}\textbf{LLCG} & \cellcolor{gray!20}$M=3$ & \cellcolor{gray!20}0.93 & 84.4 & 78.61\footnotesize{$\pm$1.23} & 84.43\footnotesize{$\pm$0.10}\cellcolor{blue!20} & \multicolumn{1}{c}{-} &  & 238.4\footnotesize{$\pm$2.3} & 184.8\footnotesize{$\pm$2.1}\cellcolor{green!20} & \multicolumn{1}{c}{-} \\
        & \cellcolor{gray!20}\textbf{\methodmultigpu} & \cellcolor{gray!20}- & \cellcolor{gray!20}1.00 & 0 & 77.75\footnotesize{$\pm$0.83} & 79.52\footnotesize{$\pm$0.24}\cellcolor{blue!20} & 47.97\footnotesize{$\pm$0.09} &  & 236.1\footnotesize{$\pm$5.5}\cellcolor{green!20} & 240.0\footnotesize{$\pm$0.0} & 177.0\footnotesize{$\pm$4.2} \\

        \bottomrule
        \end{tabular}
    }
\end{table*}
\endgroup

\begin{table*}[t]
    \centering
    \caption{Comparison of link prediction performance ($\textrm{MRR}$) and convergence time (minutes) under different base models (i.e., GCN, RGCN) and link prediction decoders (i.e., \underline{M}LP, \underline{D}istMult).
    The {convergence time is reported as the time to reach within 1\% interval of the maximum validation MRR.}
    We also report for each dataset and approach the ratio $r$ of the edges in the training graph that are available, and the preprocessing time of METIS (if needed). 
    We highlight within each row the GNN model with the best MRR in blue, and the least convergence time in green. 
    ``OOM'' denotes that experiments run out of memory.
    }
    \label{tab:exp-ablation-base-models-ecomm}
    {\small 
    \vspace{-0.3cm}
    \resizebox{\textwidth}{!}{
        \begin{tabular}{c c c c c cccc c rrrr rrr}
        \toprule
        
        \multirow{2.5}{*}{\textbf{\shortstack[c]{Dataset\\\textnormal{($|\edgeSet|$)}}}}
        & \multirow{2.5}{*}{\textbf{\shortstack[c]{Train\\Approach}}} 
        & \multirow{2.5}{*}{\textbf{\shortstack[c]{\#Parts\\($N$)}}}
        & \multirow{2.5}{*}{\textbf{\shortstack[c]{Ratio \\\textnormal{($r$)}}}} 
        & \multirow{2.5}{*}{\textbf{\shortstack[c]{Prep.\\Time\\\textnormal{\small{(mins)}}}}}
        & \multicolumn{4}{c}{\textbf{Test MRR} (\%)} &  
        & \multicolumn{4}{c}{\textbf{Conv. Time} (min)}
        \\
         \cmidrule{6-9} \cmidrule{11-14} 
        & & & & 
        & \multicolumn{1}{c}{\textbf{GCN-M}} & \multicolumn{1}{c}{\textbf{GCN-D}} & \multicolumn{1}{c}{\textbf{RGCN-M}} & \multicolumn{1}{c}{\textbf{RGCN-D}} & 
        & \multicolumn{1}{c}{\textbf{GCN-M}} & \multicolumn{1}{c}{\textbf{GCN-D}} & \multicolumn{1}{c}{\textbf{RGCN-M}} & \multicolumn{1}{c}{\textbf{RGCN-D}} \\
        \midrule

        \multirow{5}{*}{\textbf{\shortstack[c]{Ecomm\\\textcolor{gray}{\textnormal{(207M)}}}}} & \textbf{\methodrnd} & $|\vertexSet|$ & 0.33 & 0 & $\underset{\scriptstyle{\pm \textrm{0.02}}}{\textrm{84.12}}$\cellcolor{blue!20} & $\underset{\scriptstyle{\pm \textrm{0.34}}}{\textrm{79.94}}$ & $\underset{\scriptstyle{\pm \textrm{0.95}}}{\textrm{33.17}}$ & $\underset{\scriptstyle{\pm \textrm{2.13}}}{\textrm{50.73}}$ &  & $\underset{\scriptstyle{\pm \textrm{20.0}}}{\textrm{52.5}}$ & $\underset{\scriptstyle{\pm \textrm{4.3}}}{\textrm{41.4}}$\cellcolor{green!20} & $\underset{\scriptstyle{\pm \textrm{11.4}}}{\textrm{66.5}}$ & $\underset{\scriptstyle{\pm \textrm{11.5}}}{\textrm{106.9}}$ \\
        & \textbf{\method} & 15,000 & 0.76 & 6.1 & $\underset{\scriptstyle{\pm \textrm{0.45}}}{\textrm{84.44}}$\cellcolor{blue!20} & $\underset{\scriptstyle{\pm \textrm{0.08}}}{\textrm{81.53}}$ & $\underset{\scriptstyle{\pm \textrm{2.79}}}{\textrm{36.22}}$ & $\underset{\scriptstyle{\pm \textrm{3.32}}}{\textrm{52.84}}$ &  & $\underset{\scriptstyle{\pm \textrm{12.8}}}{\textrm{126.3}}$ & $\underset{\scriptstyle{\pm \textrm{17.5}}}{\textrm{105.2}}$ & $\underset{\scriptstyle{\pm \textrm{1.2}}}{\textrm{33.4}}$\cellcolor{green!20} & $\underset{\scriptstyle{\pm \textrm{12.8}}}{\textrm{132.3}}$ \\
        \noalign{\vskip 0.25ex}
        \cdashline{2-14}[0.8pt/2pt]
        \noalign{\vskip 0.25ex}
        & \cellcolor{gray!20}\textbf{\methodbasic} & \cellcolor{gray!20}$M=3$ & \cellcolor{gray!20}0.96 & 4.8 & $\underset{\scriptstyle{\pm \textrm{0.27}}}{\textrm{83.51}}$\cellcolor{blue!20} & $\underset{\scriptstyle{\pm \textrm{0.46}}}{\textrm{81.42}}$ & $\underset{\scriptstyle{\pm \textrm{3.94}}}{\textrm{38.60}}$ & $\underset{\scriptstyle{\pm \textrm{2.82}}}{\textrm{55.23}}$ &  & $\underset{\scriptstyle{\pm \textrm{39.9}}}{\textrm{121.4}}$ & $\underset{\scriptstyle{\pm \textrm{2.9}}}{\textrm{91.0}}$\cellcolor{green!20} & $\underset{\scriptstyle{\pm \textrm{11.6}}}{\textrm{188.0}}$ & $\underset{\scriptstyle{\pm \textrm{11.4}}}{\textrm{155.7}}$ \\
        & \cellcolor{gray!20}\textbf{LLCG} & \cellcolor{gray!20}$M=3$ & \cellcolor{gray!20}0.96 & 4.8 & $\underset{\scriptstyle{\pm \textrm{0.40}}}{\textrm{83.14}}$\cellcolor{blue!20} & $\underset{\scriptstyle{\pm \textrm{0.14}}}{\textrm{80.60}}$ & (OOM) & (OOM) &  & $\underset{\scriptstyle{\pm \textrm{13.4}}}{\textrm{91.4}}$ & $\underset{\scriptstyle{\pm \textrm{8.4}}}{\textrm{88.3}}$\cellcolor{green!20} & (OOM) & (OOM) \\
        & \cellcolor{gray!20}\textbf{\methodmultigpu} & \cellcolor{gray!20}- & \cellcolor{gray!20}1.00 & 0 & $\underset{\scriptstyle{\pm \textrm{0.42}}}{\textrm{82.13}}$\cellcolor{blue!20} & $\underset{\scriptstyle{\pm \textrm{1.33}}}{\textrm{81.35}}$ & (OOM) & (OOM) &  & $\underset{\scriptstyle{\pm \textrm{0.3}}}{\textrm{87.4}}$\cellcolor{green!20} & $\underset{\scriptstyle{\pm \textrm{30.8}}}{\textrm{199.3}}$ & (OOM) & (OOM) \\

        \bottomrule
        \end{tabular}
    }
    }
\end{table*}

\section{Additional Details on Experiments}
\label{app:exp-details}
\paragraph{Details on Trainer Setup} To balance the load of physical instances while keeping the empirical analysis resource- and cost-efficient, 
we implement the following trainer setup: 
For the largest dataset, MAG240M-P, with $M=3$ trainers, we run the \maframework{} server and one trainer on physical instance \#1 and the other two trainers on instance \#2. For $M=5$ trainers, we run the two additional trainers on instance \#3. 
For the smaller datasets (Reddit, ogbl-citation2 and E-comm), we run the \maframework{} server and $M=3$ or $M=5$ trainers on a single physical instance. For all experiments with $M=23$ trainers, we use all 24 GPUs on three physical instances, with one GPU reserved for model evaluation on the server process. 

\vspace{0.1cm}
\paragraph{Hyperparameter Choices} We tune and select the best-performing hyperparameters on the GGS baseline, and adopt the same hyperparameters for distributed training approaches for a fair comparison. 
\begin{itemize}
    \item For Reddit and ogbl-citation2, we use 2-layer models with the size of hidden representations as 256 for both the encoder and decoder;
    \item For the larger MAG240M-P dataset, we use 2-layer models with the size of hidden representations as 64. 
    \item For E-comm, we use 2-layer GCN or RGCN models as the encoder, and DistMult or 2-layer MLP as the decoder. For GCN, we set the dimension of hidden representations as 128. To reduce the memory usage of RGCN, we adopt basis decomposition~\cite{schlichtkrull2018modeling} with 4 bases (equal to the total number of forward and inverse relations), each with 128 dimension, and added an MLP layer before RGCN input to reduce the dimension of input representations to 128. For DistMult decoder, we set the dimension of each relational embedding as 128. For MLP decoder, we use 2 layers with the size of hidden representations as 128.
\end{itemize} 
We set the learning rate $\texttt{lr}=0.001$ in all the experiments, since we find that it significantly improves the performance compared to $\texttt{lr}=0.01$. 
For all experiments, we allocate 4-hour training time; in most cases, this is sufficient time for models to reach convergence (as shown in Fig.~\ref{fig:exp-benchmark-MRR}), while not incurring excessive time and monetary cost.

\paragraph{Formulation of MLP Decoder} Given the embeddings $\mathbf{r}_u$ and $\mathbf{r}_v$ for nodes $u, v$ by GNN encoder, respectively, the $k$-th layer of the MLP decoder is formulated as 
$\mathbf{e}_{u,v}^{(k+1)} = \sigma(\mathbf{e}_{u,v}^{(k)}\mathbf{\Theta}^{(k)})$, where $\mathbf{\Theta}^{(k)}$ is the learnable weight matrix, and $\mathbf{e}_{u,v}^{(0)} = \mathbf{r}_u \odot \mathbf{r}_v$ is the element-wise product of $\mathbf{r}_u$ and $\mathbf{r}_v$; 
the predicted link probability $\hat{y}_{u,v} = \mathbf{e}_{u,v}^{(K)}$ equals to the output scalar for a $K$-layer decoder. 
We adopt PReLU as the activation function $\sigma$, as we do for the encoders.

\vspace{0.1cm}
\paragraph{Ablation on Base Models}
We list the performance and convergence time of the training approaches on different base models for homogeneous datasets (i.e., GCN~\cite{kipf2016semi}, GraphSAGE~\cite{hamilton2017inductive} and MLP encoders; all with MLP decoder) in Table~\ref{tab:exp-ablation-base-models}, with results on base models for E-comm dataset (i.e., GCN~\cite{kipf2016semi} and RGCN~\cite{schlichtkrull2018modeling} with MLP or DistMult~\cite{yang2015embedding} decoder) in Table~\ref{tab:exp-ablation-base-models-ecomm}; in Table~\ref{tab:exp-benchmark} of the main paper, we report the results for the best-performing base model per approach and dataset. 
We did not test MLP for \llcgbasic{}, as MLP is graph-agnostic and does not benefit from the \llcgbasic{} global model correction process for recovering cross-partition edges~\cite{ramezani2021learn}. 

On homogeneous datasets (Table~\ref{tab:exp-ablation-base-models}), we observe that GCN and GraphSAGE are the best-performing base models for all approaches on Reddit and MAG240M-P, respectively; on ogbl-citation2, the best-performing base models vary for different training approaches.

On the heterogeneous E-comm dataset (Table~\ref{tab:exp-ablation-base-models-ecomm}), we surprisingly observe that GCN, which ignores the heterogeneous edge types in the dataset, outperforms RGCN designed for heterogeneous graphs by a large margin.
Prior works have also observed that modeling heterogeneous relations in GNN models may not be as crucial as widely presumed~\cite{zhang2022rethinking,li2022graph}, and we leave for the future works for further investigation of this finding.

\section{Proofs of Theorems}
\label{app:proofs}

\begin{proof}[\textbf{Proof for Lemma~\ref{thm:homophily-min-cut}}]
\label{prf:homophily-min-cut}
    Following the assumption, the probability $p_{ji}$ of node $j$ to connect to node $i$ can be written as
    \begin{equation*}
        p_{ji} = \frac{\mathbf{H}(y_i, y_j)}{\sum_{l\in \vertexSet} \mathbf{H}(y_l, y_j)} = \frac{1}{C} \mathbf{H}(y_i, y_j) = 
        \begin{cases}
            h / C, & \text{if } y_i = y_j \\
            (1-h) / C, & \text{if } y_i \neq y_j
        \end{cases}
        \vspace{-0.2cm}
    \end{equation*}
    Now assume that the ratio of nodes $v$ with label $y_v = 0$ in partition 1 as $\beta_1$, and in partition 2 as $\beta_2$, then we have the feature distributions $\mathbf{C}_1$ and $\mathbf{C}_2$ in each partition, which follow the distributions for class labels $0$ and $1$ under onehot-encoded node features $\V{x}_v$, as $\mathbf{C}_1 = [\beta_1, 1 - \beta_1]$ and $\mathbf{C}_2 = [\beta_2, 1 - \beta_2]$. Since we assume that the two partitions have equal sizes $\eta$, and two class labels with equal sizes, we have $\beta_1 \eta + \beta_2 \eta = (1 - \beta_1) \eta + (1 - \beta_2) \eta$ and $\beta_2 = 1 - \beta_1$. 
    Thus, we denote $\beta_1 = \beta \in [0, 1]$ and simplify $\mathbf{C}_1$ and $\mathbf{C}_2$ as $\mathbf{C}_1 = [\beta, 1 - \beta]$ and $\mathbf{C}_2 = [1 - \beta, \beta]$; without loss of generality, we assume $\beta \geq 0.5$. 

    We denote the random variable $A_{ij} = 1$ if an edge exists between node $i$ and $j$, and $A_{ij} = 0$ otherwise. Then we have $\expect{A_{ij}} = p_{ji}$. 
    The expected number of edge cuts between the two partitions $\lambda$ is
    \begin{equation}
        \lambda = \sum_{\mathclap{\substack{i \in \alpha^{-1}(1)\\j \in \alpha^{-1}(2)}}} \expect{A_{ij}} = \sum_{\mathclap{\substack{i \in \alpha^{-1}(1), y_i = 0\\j \in \alpha^{-1}(2)}}} \expect{A_{ij}} \;+\;\sum_{\mathclap{\substack{i \in \alpha^{-1}(1), y_i = 1\\j \in \alpha^{-1}(2)}}} \expect{A_{ij}} \coloneqq 
        (\lambda_0 + \lambda_1) / C
    \end{equation}
    and it is straightforward to show that $\lambda_0 = \beta \eta ((1-\beta) \eta h + \beta \eta (1-h))$ and $\lambda_1 = (1 - \beta) \eta ((1-\beta) \eta (1 - h) + \beta \eta h)$. As a result, we have
    \begin{equation}
        \vspace{-0.1cm}
        \lambda = \left(1 - 2 (1 - \beta) \beta -(2 \beta - 1)^2 h\right) \frac{\eta ^2}{C}
    \end{equation}
    For $\beta \in [0.5, 1]$ and homophilic graph with $h \geq 0.5$, it is easy to show that $\lambda$ reaches the minimal value when $\beta = 1$, and $\mathbf{C}_1 = [1, 0]$ and $\mathbf{C}_2 = [0, 1]$. Therefore, we show that the minimal expected edge cut is reached when each partition contains only the node from a single class with the same features.
\end{proof}

\begin{proof}[\textbf{Proof for Theorem~\ref{thm:homophily-grad-loss}}]
\label{prf:homophily-grad-loss}
    We note that under the L2-loss function 
    $\mathcal{L}(y, z) = \tfrac{1}{2}\Vert \V{y} - \V{z} \Vert^2$, the loss value (or gradient) for a training batch with multiple nodes is the sum of the loss value (or gradient) calculated individually on each node; thus, we can simplify our discussion by only examining the loss value and gradient for a single node. For an arbitrary node $v \in \vertexSet$ for training, we have
    \begin{equation}
        \label{eq:proof-2-pred}
        z_v = \sigma\left( \sum_{u \in \neighbors(v)} \frac{1}{d_v} \mathbf{x}_u^\T\mathbf{w}\right) \coloneqq \sigma(g(\V{w})) = \frac{1}{1 + e^{-g(\V{w})}}
    \end{equation}
    Without loss of generality, we assume the class label of node $v$ as $y_v = 1$. 
    In this case, we have the loss function $\mathcal{L}(y_v, z_v)$ as 
    \begin{equation}
        \label{eq:proof-2-loss}
        \mathcal{L}(y_v, z_v) = \tfrac{1}{2}(\sigma(g(\V{w})) - 1)^2.
    \end{equation}
    \paragraph{Discrepancies Among Expected Initial Gradients} The gradient of the model weights $\V{w}$ for training node $v$ is
    \begin{equation}
        \label{eq:proof-2-grad}
        \nabla\loss = \frac{\partial L}{\partial \V{w}} = (\sigma(g(\V{w})) - 1) \cdot \frac{\partial \sigma(g(\V{w}))}{\partial g(\V{w})} \cdot \frac{\partial g(\V{w})}{\partial \V{w}}.
    \end{equation}
    As $\sigma$ is the sigmoid function, we have 
    \begin{equation}
        \label{eq:proof-2-grad-act}
       \frac{\partial \sigma(g(\V{w}))}{\partial g(\V{w})} = \sigma(g(\V{w})) (1 - \sigma(g(\V{w})))
    \end{equation}
    \begin{equation}
        \label{eq:proof-2-grad-agg}
        \frac{\partial g(\V{w})}{\partial \V{w}} = \sum_{u \in \neighbors(v)} \frac{1}{d_v} \mathbf{x}_u
     \end{equation}
    Now we look into how the gradients change when we ignore the cross-partition edges under model aggregation training, which changes the \emph{effective} neighborhood $\neighbors'(v)$ of node $v$ in Eq.~\eqref{eq:proof-2-pred} and \eqref{eq:proof-2-grad-agg}. 
    Following the analyses in Proof~\hyperref[prf:homophily-min-cut]{1}, we can assume the feature distribution $\mathbf{C}_1$ and $\mathbf{C}_2$ in each partition as $\mathbf{C}_1 = [\beta, 1 - \beta]$ and $\mathbf{C}_2 = [1 - \beta, \beta]$, where $\beta \in [0, 1]$; the difference of the group distributions $\Vert \mathbf{C}_{2} - \mathbf{C}_{1} \Vert = \sqrt{2} | 1-2 \beta |$.

    \textcircled{\raisebox{-0.9pt}{1}} For \emph{centralized training}, the effective neighborhood $\neighbors'(v)$ of node $v$ is equal to its actual neighborhood $\neighbors(v)$. Thus, for $y_v=1$, and the assumed node features $\V{x}_v = \mathrm{onehot}(y_v)$, we have
    \begin{equation}
        \label{eq:proof-2-exp-agg-central}
        \expectlarge{\sum_{u \in \neighbors(v)} \frac{1}{d_v} \mathbf{x}_u} 
        = \frac{1}{d_v}
        \begin{bmatrix}
            (1-h) d_v & h d_v
        \end{bmatrix}
        = \begin{bmatrix}
            (1-h)  & h
        \end{bmatrix}
    \end{equation}
    When initializing $\V{w} = 0$, we have $g(\V{w}) = 0$ and $\sigma(g(\V{w})) = 0.5$. Combining Eq.~\eqref{eq:proof-2-grad}-\eqref{eq:proof-2-exp-agg-central}, we have the expected initial gradient $\expect{\nabla\loss^{global}}$ for centralized training as
    \begin{equation}
        \label{eq:proof-2-exp-grad-central}
        \expect{\nabla\loss^{global}} = -\tfrac{1}{8}\begin{bmatrix}
            (1-h)  & h
        \end{bmatrix}
    \end{equation}

    \textcircled{\raisebox{-0.9pt}{2}} When $v$ is on \emph{instance 1} with class distribution $\mathbf{C}_1 = [\beta, 1 - \beta]$, the effective neighborhood $\neighbors'(v)$ of node $v$ is changed compare to its actual neighborhood. In this case, we have for $y_v=1$
    \begin{equation}
        \label{eq:proof-2-exp-agg-1}
        \expectlarge{\frac{\partial g(\V{w})}{\partial \V{w}}} 
        = \frac{1}{d_v((1-h)\beta + h(1-\beta))}
        \begin{bmatrix}
            (1-h)\beta d_v & h (1-\beta) d_v
        \end{bmatrix},
    \end{equation}
    and when initializing $\V{w} = 0$, we have the expected initial local gradient $\expect{\nabla\loss^{local}_1}$ on instance 1 as
    \begin{equation}
        \label{eq:proof-2-exp-grad-1}
        \expect{\nabla\loss^{local}_1} = 
        -\frac{1}{8((1-h)\beta + h(1-\beta))}
        \begin{bmatrix}
            (1-h)\beta  & h (1-\beta) 
        \end{bmatrix}.
    \end{equation}

    \textcircled{\raisebox{-0.9pt}{3}} When $v$ is on \emph{instance 2} with class distribution $\mathbf{C}_1 = [1 - \beta, \beta]$, we have for $y_v=1$
    \begin{equation}
        \label{eq:proof-2-exp-agg-2}
        \expectlarge{\frac{\partial g(\V{w})}{\partial \V{w}}} 
        = \frac{1}{d_v((1-h)(1-\beta) + h\beta)}
        \begin{bmatrix}
            (1-h)(1-\beta) d_v & h \beta d_v
        \end{bmatrix}
    \end{equation}
    And when initializing $\V{w} = 0$, we have the expected initial local gradient $\expect{\nabla\loss^{local}_2}$ on instance 2 as
    \begin{equation}
        \label{eq:proof-2-exp-grad-2}
        \expect{\nabla\loss^{local}_2} = 
        -\frac{1}{8((1-h)(1-\beta) + h\beta)}
        \begin{bmatrix}
            (1-h)(1-\beta) d_v & h \beta d_v
        \end{bmatrix}
    \end{equation}

    Based on Eq.~\eqref{eq:proof-2-exp-grad-central}, \eqref{eq:proof-2-exp-grad-1}, \eqref{eq:proof-2-exp-grad-2}, we have the discrepancies measured under $l^2$-norm between these expected initial gradients as
    \begin{align*}
        \Vert \expect{\nabla\loss^{global}} - \expect{\nabla\loss^{local}_1} \Vert_2 & = \frac{\sqrt{2}}{8}\left| \frac{(1-2 \beta ) (h-1) h}{\beta -2 \beta  h+h}\right| \\
        \Vert \expect{\nabla\loss^{global}} - \expect{\nabla\loss^{local}_2} \Vert_2 & = \frac{\sqrt{2}}{8}\left| \frac{(2 \beta -1) (h-1) h}{1 - \beta + (2 \beta -1) h} \right| \\
        \Vert \expect{\nabla\loss^{local}_1} - \expect{\nabla\loss^{local}_2} \Vert_2 & = \left| \frac{\tfrac{1}{4 \sqrt{2}}(2 \beta -1) (h-1) h}{(\beta -2 \beta  h+h-1) (\beta -2 \beta  h+h)} \right|
    \end{align*}
    Given that the difference of the group distributions $\Vert \mathbf{C}_{2} - \mathbf{C}_{1} \Vert = \sqrt{2} | 1-2 \beta |$, it is straightforward to see from the above equations that (1) there are no discrepancy among all initial gradients when $\beta = 0.5$, and (2) the discrepancies increase with the increase of $\Vert \mathbf{C}_{2} - \mathbf{C}_{1} \Vert = \sqrt{2} | 1-2 \beta |$ when $h \geq 0.5$.

    \vspace{0.1cm}
    \paragraph{Discrepancies Among Expected Loss Values} We only show the proof for node $v$ with class label $y_v=1$ here; the case of $y_v=0$ can be proved in a similar way. Assume the model weight $\V{w} = [w_0, w_1]$; based on Eq.~\eqref{eq:proof-2-pred}, \eqref{eq:proof-2-loss}, \eqref{eq:proof-2-grad-agg}, \eqref{eq:proof-2-exp-agg-1}, and \eqref{eq:proof-2-exp-agg-2}, we have for instance 1 and instance 2, when not considering cross-partition edges,
    \begin{equation*}
        \expect{\loss_{1}^{local}(\matW)} = \left(1 + 
        \mathrm{exp}\left(
            \frac{\beta  (h-1) w_0+(\beta -1) h w_1}{(2 \beta -1) h-\beta }
            \right) \right)^{-2},
    \end{equation*}
    \begin{equation*}
        \expect{\loss_{2}^{local}(\matW)} = \left(1 + 
        \mathrm{exp}\left(
            \frac{(\beta -1) (h-1) w_0+\beta  h w_1}{-\beta +(2 \beta -1) h+1}
            \right) \right)^{-2}.
    \end{equation*}
    Note that function $(1+\mathrm{exp}(x))^{-2}$ monotonically decreases with variable $x$, therefore $\expect{\loss_{1}^{local}(\matW)} = \expect{\loss_{2}^{local}(\matW)}$ if and only if
    \begin{equation}
        \label{eq:proof-2-loss-equal-cond}
        \frac{\beta  (h-1) w_0+(\beta -1) h w_1}{(2 \beta -1) h-\beta } = \frac{(\beta -1) (h-1) w_0+\beta  h w_1}{-\beta +(2 \beta -1) h+1}.
    \end{equation}
    Eq.~\eqref{eq:proof-2-loss-equal-cond} holds if and only if $\beta = 0.5$, which means $\Vert \mathbf{C}_{2} - \mathbf{C}_{1} \Vert = \sqrt{2} | 1-2 \beta | = 0$. Therefore, the expected loss values $\expect{\loss_{i}^{local}(\matW)}$ on each instance $i\in\{1,2\}$, without considering cross-partition edges, is equal if and only if $\mathbf{C}_{1} = \mathbf{C}_{2}$.
\end{proof}

\end{document}